\documentclass[10pt,journal,compsoc]{IEEEtran}

\ifCLASSOPTIONcompsoc
\usepackage[nocompress]{cite}
\usepackage{mathrsfs}
\usepackage{amssymb}
\usepackage{algorithm}
\usepackage{amsmath}
\usepackage{amsthm}
\usepackage{algpseudocode}
\usepackage{array}
\usepackage{booktabs} % 三线表
\usepackage{bm}
\usepackage{color}
\usepackage{colortbl}
\usepackage{enumitem}
\usepackage{epsfig}
\usepackage{float}
\usepackage{graphicx}
\usepackage{lineno}
\usepackage{setspace}
\usepackage{indentfirst}
\usepackage{multirow}
\usepackage{algorithm}
\usepackage{microtype}
\usepackage{graphicx}
\usepackage{subfigure}
\usepackage{booktabs} % for professional tables
\usepackage{amssymb}
\usepackage{enumerate}
\usepackage{mathrsfs}
\usepackage{graphicx}
\usepackage{float}
\usepackage{algorithm}
\usepackage{amsmath}
\usepackage{amsthm}
\usepackage{amssymb}
\usepackage{lineno}
\usepackage{setspace}
\usepackage{indentfirst}
\usepackage{epsfig}
\usepackage{multirow}
\usepackage{bm}
\usepackage{makecell} % 替换三线表
\usepackage{color}
\usepackage{colortbl}
\usepackage{hyperref}
\usepackage{lineno}
\usepackage{amssymb}
\usepackage{mathrsfs}
\usepackage{graphicx}
\usepackage{float}
\usepackage{algorithm}
\usepackage{amsthm}
\usepackage{amssymb}
\usepackage{setspace}
\usepackage{indentfirst}
\usepackage{epsfig}
\usepackage{multirow}
\usepackage{bm}
\usepackage[ragged]{footmisc}
\usepackage{scrextend}

\usepackage{amsmath}

\else
  \usepackage{cite}
\fi
\ifCLASSINFOpdf
\else
\fi
\begin{document}
\title{Adaptive Neighborhood Metric Learning}
\author{Kun Song, Junwei Han,~\IEEEmembership{Senior Member,~IEEE}, Gong Cheng,
  Jiwen Lu,~\IEEEmembership{Senior Member,~IEEE}, Feiping Nie% <-this % stops a space
\IEEEcompsocitemizethanks{
\IEEEcompsocthanksitem Kun Song, Junwei Han (corresponding author) and Gong Cheng are with School of Automation, Northwestern Polytechnical University, China.\protect\\
E-mail: \{songkun123000, junweihan2010, chenggong1119\}@gmail.com
\IEEEcompsocthanksitem Kun Song is also with the information Engineering College of Yangzhou University.
\IEEEcompsocthanksitem Jiwen Lu is with the State Key Lab of Intelligent Technologies and Systems, Beijing National Research Center for Information Science
and Technology (BNRist), Beijing 100084, China, and also with the Department of Automation, Tsinghua University, Beijing 100084, China.\protect\\
 E-mail: lujiwen@tsinghua.edu.cn.
\IEEEcompsocthanksitem
Feiping Nie is with School of Computer Science and Center for Optical Imagery Analysis and Learning (OPTIMAL), Northwestern Polytechnical University, Xi'an 710072, Shaanxi P.R. China.\protect\\
E-mail: feipingnie@gmail.com% <-this % stops an unwanted space
 }% <-this % stops an unwanted space
}
\markboth{Journal of \LaTeX\ Class Files,~Vol.~14, No.~8, August~2015}%
{Shell \MakeLowercase{\textit{et al.}}: Bare Demo of IEEEtran.cls for Computer Society Journals}

\IEEEtitleabstractindextext{%
\begin{abstract}
 In this paper, we reveal that metric learning would suffer from serious inseparable problem if without informative sample mining. Since the inseparable samples are often mixed with hard samples, current informative sample mining strategies used to deal with inseparable problem may bring up some side-effects, such as instability of objective function, etc. To alleviate this problem, we propose a novel distance metric learning algorithm, named adaptive neighborhood metric learning (ANML). In ANML, we design two thresholds to adaptively identify the inseparable similar and dissimilar samples in the training procedure, thus inseparable sample removing and metric parameter learning are implemented in the same procedure. Due to the non-continuity of the proposed ANML, we develop an ingenious function, named \emph{log-exp mean function} to construct a continuous formulation to surrogate it, which can be efficiently solved by the gradient descent method. Similar to Triplet loss, ANML can be used to learn both the linear and deep embeddings. By analyzing the proposed method, we find it has some interesting properties. For example, when ANML is used to learn the linear embedding, current famous metric learning algorithms such as the large margin nearest neighbor (LMNN) and neighbourhood components analysis (NCA) are the special cases of the proposed ANML by setting the parameters different values. When it is used to learn deep features, the state-of-the-art deep metric learning algorithms such as Triplet loss, Lifted structure loss, and Multi-similarity loss become the special cases of ANML. Furthermore, the \emph{log-exp mean function} proposed in our method gives a new perspective to review the deep metric learning methods such as Prox-NCA and N-pairs loss. At last, promising experimental results demonstrate the effectiveness of the proposed method.
  %The code is available at \href{}{https://github.com/huanwoheshan/anml.git}
\end{abstract}
}
\maketitle
\begin{figure*}[t]
 \centering
   \includegraphics[width=0.85\linewidth]{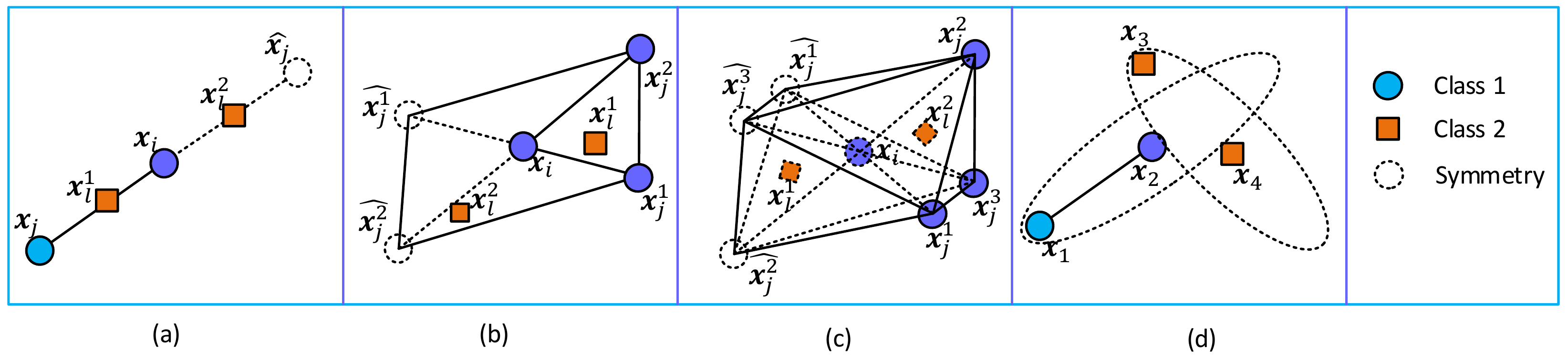}
    \vspace{-7pt}\caption{The illustration of the geometric structures (inseparable regions) to cause the inseparable samples. In (a)-(c), the query is from class 1, the dissimilar samples are from class 2. There is no metric value to let dissimilar samples out of line segment (a), parallelogram (b), or polyhedron (c) under linear metric learning. In (d), there is no metric value to let $(\textbf{x}_2,\textbf{x}_1,\textbf{x}_4)$ and $(\textbf{x}_4,\textbf{x}_3,\textbf{x}_2)$ be separable simultaneously.}
  \label{fig_neb_r_line}
\end{figure*}

\IEEEdisplaynontitleabstractindextext
\IEEEpeerreviewmaketitle
\IEEEraisesectionheading{\section{Introduction}\label{sec:introduction}}
 \indent Distance metric learning is a fundamental machine learning topic which possesses enormously wide spectrum of applications, such as feature reduction \cite{davis2007information,wang2015survey,Wu2016Online}, recognition \cite{yen2016pd,song2017parameter,Ying2018Manifold}, person re-identification \cite{cheng2016person}, visual tracking \cite{7532650,li2012non}, and image classification\cite{Qian2014An,NIPS2012_4808,6236161}, etc. Commonly, metric learning aims at learning a proper projection function to pull similar samples close and simultaneously push dissimilar samples apart from each other, which can be described by a set of triplets in which one triplet $(\textbf{x}_i,\textbf{x}_j,\textbf{x}_l)$ implies a triplet-based discriminant criterion: the distance between the similar pair $(\textbf{x}_i, \textbf{x}_j)$ should be smaller than that between the dissimilar pair $(\textbf{x}_i,\textbf{x}_l)$ after projection. Thus, metric learning is a typical type of discriminant analysis.\\
  \indent In this paper, we find the triplet-based discriminant criterion easily encounters inseparable problem if without informative sample mining, which would make the model converge at a local optimal early. According to the finding, we provide a new understanding of informative sample mining in metric learning, i.e., to remove inseparable samples. This gives a more convincing explanation of some problems in informative sample mining, such as the contradiction between semi-hard sample mining \cite{Schroff2015FaceNet} (or weighted sample mining \cite{wu2017sampling}) and hard sample mining. At last, we propose an effective method to solve the inseparable problem according to our finding.\\
   \indent We start from liner metric learning to demonstrate the inseparable problem of metric learning. As seen in Fig.\ref{fig_neb_r_line}(a), when the dissimilar sample $\textbf{x}_l^1$ lies on the line segment formed by similar samples $\textbf{x}_i$ and $\textbf{x}_j$, the triplet $(\textbf{x}_i,\textbf{x}_j,\textbf{x}^1_l)$ will never satisfy the triplet-based discriminant criterion, and become an inseparable triplet. That is because the geometrical structure of the line segment is preserved by the linear projection, i.e., those three samples would still on a line segment after projection. Moreover, the geometry structure will be amplified by the symmetry of distance function. Look back at Fig. \ref{fig_neb_r_line}(a), when the dissimilar sample $\textbf{x}_l^2$ lies on the reverse extension of $\textbf{x}_i\textbf{x}_j$, i.e., the line segment $\textbf{x}_i\hat{\textbf{x}_j}$, the triplet $(\textbf{x}_i,\textbf{x}_j,\textbf{x}^2_l)$ is still inseparable. This indicates that two classes of samples never intersecting with each other, can still produce many inseparable triplets. Besides line segments, many other types of geometrical structures are also preserved by linear projection, such as parallelograms and polyhedrons. The illustrations are shown in Fig.\ref{fig_neb_r_line}(b)-(c). For convenience, we call those structures as inseparable regions, and we have mathematically proved that their shapes depend on the number of similar samples of each query. Thus, more similar samples construct higher dimensional inseparable region, which may produce more inseparable dissimilar samples.\\
 \indent Metric learning algorithms always depict the discriminant information of different classes by one same learned parameter, which would enlarge inseparable regions of queries. As shown in Fig.\ref{fig_neb_r_line}(d), one of two inseparable triplets $(\textbf{x}_2,\textbf{x}_1,\textbf{x}_4)$ and $(\textbf{x}_4,\textbf{x}_3,\textbf{x}_2)$ would become inseparable due to having different feasible solutions. Suppose $(\textbf{x}_2,\textbf{x}_1,\textbf{x}_4)$ is separable, all the dissimilar samples of $\textbf{x}_2$ located on the line determined by $\textbf{x}_4$ and $\textbf{x}_3$ are inseparable. Considering the original inseparable regions of the two triplets, i.e., line segments $\textbf{x}_2\textbf{x}_1$, $\textbf{x}_4\textbf{x}_3$, and their inverse line segments, thus the whole line $\textbf{x}_1\textbf{x}_2$ except the line segment $\textbf{x}_1\textbf{x}_2$ is the enlarged part. When the class number is larger, the searching spaces of queries from different classes would be contradicted more seriously, and more enlarged inseparable regions are obtained.\\
 \indent Since so many factors tend to complicate inseparable regions, a simple dataset may construct very complex inseparable regions and have lots of inseparable samples. Due to the extraordinary complexity of the inseparable regions, those training samples may not support a trained non-linear projections that is powerful enough to pull the inseparable samples out of them. Since the dimension and enlargement of inseparable regions are all increased with the amount of training samples, the situation would be not improved even if the training set is large. That is because there is an endless loop: More training samples produce more complex inseparable regions which need more powerful non-linear projection to project dissimilar samples out of them, and inversely needs more training samples to learn the large amount of parameters. The endless loop indicates that the inseparable problem of metric learning could not be effectively solved by the non-linear projection. This also means non-linear metric learning suffers from serious inseparable problem.\\
\indent Generally, for each query sample, its nearest dissimilar samples and the farthest similar samples under the learned metric are most likely inseparable. Thus, the inseparable triplets could be reduced by removing those nearest dissimilar and farthest similar samples. This procedure is very similar to the informative sample mining, especially the semi-hard sample mining. Thus, we argue that informative sample mining acts the role to remove the inseparable samples. However, the inseparable samples are often mixed with the hard samples which produce gradient with large magnitudes. This makes sample mining strategies often come into being some side-effects. For example, the semi-hard sample mining will lead to the instability of objective function since it removes the hardest samples, while the hard sample mining preserves the most inseparable samples since the hardest samples are likely inseparable. Those side-effects would hurt the performance of metric learning.\\
 \indent To alleviate the inseparable problem, we propose a novel metric learning method named adaptive neighborhood metric learning. In the proposed algorithms, two thresholds are designed to adaptively distinct the separable similar and dissimilar samples, which are functions of the number of separable similar (or dissimilar) samples. By using those thresholds, the separable samples can be automatically found during the training procedure, meanwhile, the metric parameter is learned to separate those separable samples. Similar to the Triplet loss which can be applied to learn linear embedding and deep embedding, our proposed loss function can also be applied to these two different learning tasks. By analyzing our method, we find it has the following properties:
\begin{itemize}
 \item By adopting linear projection, our method becomes linear adaptive neighborhood metric learning (LANML). We have proved that the famous methods large margin nearest neighbor (LMNN)\cite{weinberger2005distance} and neighbourhood components analysis (NCA) \cite{Goldberger2004Neighbourhood} are both the special cases of LANML. Therefore, our work bridges a connection between the convex metric learning model LMNN and the non-convex model NCA.
  \item By adopting deep neural networks to project the samples, our method becomes deep adaptive neighborhood metric learning (DANML). We find that the DANML is the general form of some state-of-the-art deep metric learning methods, such as the FaceNet, the Lifted structure loss, and the Multi-similarity loss, etc.
  \item With the concept of separable sample selection and our proposed \emph{log-exp mean function}, we give an intuitive perspective to review the existing state-of-the-art methods, such as the Proxy-NCA and N-pairs loss, and give a simple explanation for why they are so effective.
\end{itemize}
\indent At last, we evaluate the proposed methods by conducting extensive experiments of classification tasks on several data sets, including the UCI datasets, the fine-grained classification datasets, and large scale image retrieval datasets. The promising results have demonstrated the superiority of the proposed methods.
\vspace{-5pt}\section{Related Work}
\subsection{Metric Learning}
 \indent Current metric learning algorithms can mainly be divided into two categories: conventional metric learning and deep metric learning. The former one learns the Mahalanobis distance\cite{song2017parameter,Di2017Large,Liu2015Large,davis2007information} or kernel-based metric learning\cite{kedem2012non,perrot2015regressive} to measure similarity between samples, while the latter one learns the CNN features based on the metric learning objective function. Generally, those two kinds of methods share the same learning paradigm to form the objective function due to their same  purposes, i.e., to pull similar samples close while push dissimilar samples apart from each other.\\
 \indent Contrastive loss\cite{davis2007information} and triplet loss\cite{Schroff2015FaceNet,Di2017Large} are two types of widely used objective functions in metric learning. In contrastive loss, similar samples and dissimilar samples are separated by a fixed margin, which makes its geometrical meaning is quit distinct. However, since contrastive loss does not allow the intra-class variances, it has bad generalization ability. As a contrast, triplet loss is more flexible due to its mere requirement of a certain ranking within triplets, yet it has more constraints considered than contrastive loss.\\
 \indent Inspired with those two learning paradigms, many novel loss functions of metric learning have been proposed. For example, Song et al.\cite{oh2016deep} proposed a lifted structured loss to consider all the positive and negative pairs within a batch. Sohn et al. \cite{sohn2016improved} designed the N-pair loss to push away N-1 negative samples in one (N+1)-tuple altogether. Yair et al.\cite{movshovitz2017no} designed a proxy-based loss to speed the training of NCA loss. Then, Kim et al. \cite{kim2020proxy} improved the proxy-based strategy to form the proxy anchor loss. Besides, there are some works developing novel sample mining strategies to enhance the triplet loss and contrastive loss \cite{wu2017sampling,ge2018deep}. More recently, some works \cite{wang2019multi,wang2020ranked} attempt to construct general frameworks to sum up those existing metric learning methods. However, most of them are derived from the computation of gradients which can not reveal the intrinsic characters of metric learning. For example, they can not give a distinct geometric explanation of the obtained objective functions. Another trend is to incorporate ensemble technique in metric learning which integrates several diverse embeddings to constitute a more informative representation \cite{kim2018attention,opitz2018deep,Shen2009Positive}.
   \vspace{-10pt}\subsection{Informative Sample Mining}
   \indent As well known, the gradients of a metric learning algorithm are often computed on a set of sample groups such as triplets, pairs, N-pairs, and qudraplets. Due to extremely large number of those sample groups, the informative sample mining is often required to reduce the redundant ones among them. For example, \cite{weinberger2005distance} selected the $K$ nearest neighbors to construct the objective function of LMNN by Euclidean distance. \cite{Schroff2015FaceNet} proposed a 'semi-hard' negative sample mining method to reduce triplets in FaceNet. \cite{dfense_triplet} proposed an online hard example mining (OHEM) algorithm to train region-based object detectors. \cite{wu2017sampling} showed the sample mining is very important to the training of metric learning by proposing a weighted sampling with margin-based loss. \cite{harwood2017smart} proposed a smart mining procedure to improve the efficiency of semi-hard sample mining. \cite{zheng2019hardness} presented a hard-aware deeply cascaded embedding approach to address the problem that hard samples are too less.\\
  \indent Although those multifarious sampling methods boosted the performance of metric learning, what the true role of informative sample mining playing in the metric learning is not clear. For example, how the redundant sample groups make the metric learning stop at an early converging point. Besides, why the sampling criteria of hard sample mining \cite{wu2017sampling} and semi-hard sample mining \cite{Schroff2015FaceNet}\cite{harwood2017smart} (or weighted sampling strategy \cite{wu2017sampling}) are opposite, which goes against our intuition that the informative samples considered by metric learning methods serving for the same purpose should be roughly the same. The unclear role of informative sample mining would hinder us to design more powerful metric learning algorithms.\\
 \indent In this paper, we try to explain the roles of informative sample mining from the perspective of inseparable problem which is a common phenomenon appeared in the discriminant analysis but rarely considered in the field of metric learning.
 \vspace{-5pt}\section{Inseparable Problem of Metric Learning}
 \vspace{-2pt}
 \subsection{How the Inseparable Samples are Produced?}
 \label{htisap_1}
%\begin{figure}[t]
% \centering
%  \includegraphics[width=0.9\linewidth]{inseparable_region.pdf}
%  \caption{The illustration of the original space and the projected space for distance metric learning. In the original space, the points in a color curve would be projected into the region in the same colored circle.} \label{fig_neb_r_pro}
%\end{figure}
\indent Suppose $\mathcal{X}=\{(\textbf{x}_i,y_i)\}_{i=1}^N$ is a set of $N$ labeled samples, where $y_i \in \{1,2,\cdots,C\}$ is the corresponding label of $\textbf{x}_i \in \mathbb{R}^d$. The projection function $\textbf{z}_i=f_\theta(\textbf{x}_i):\mathbb{R}^d\mapsto\mathbb{R}^d$ parameterized with $\theta$ derives a distance function on two samples $\textbf{x}_i$ and $\textbf{x}_j$ denoted as $d_\theta(\textbf{x}_i,\textbf{x}_j) = \vert \textbf{z}_i-\textbf{z}_j\vert$. For each query sample $\textbf{x}_i$, its similar set $\mathcal{S}_i$ consists of similar samples whose labels are $y_i$ and the corresponding dissimilar set $\mathcal{D}_i$ consists of dissimilar samples whose labels are different from $y_i$. According to the triplet constraints \cite{Schroff2015FaceNet,weinberger2005distance}, the purpose of metric learning is to find a suitable $\theta$ to let the dissimilar samples in $\mathcal{D}_i$ out of the neighborhood:
\begin{equation}\label{sim_neighbor}
\small
\mathcal{A}_i(\theta)=\bigcup_{j\in \mathcal{S}_i}\{\textbf{x} | d_\theta(\textbf{x}_i,\textbf{x}) < d_\theta(\textbf{x}_i,\textbf{x}_j)\},
\end{equation}
or to let similar samples in $\mathcal{S}_i$ into the neighborhood:
 \begin{equation}\label{dissim_neighbor}
 \small
 \mathcal{B}_i(\theta)=\bigcap_{l\in \mathcal{D}_i}\{\textbf{x} | d_\theta(\textbf{x}_i,\textbf{x}) < d_\theta(\textbf{x}_i,\textbf{x}_l)\}
 \end{equation}
\indent In this way, the similar samples in $\mathcal{S}_i$ are separated from the dissimilar samples in $\mathcal{D}_i$ for each query $\textbf{x}_i$ by the boundary of $\mathcal{A}_i(\theta)$ (or $\mathcal{B}_i(\theta)$). To explain how the inseparable samples are produced, we define the inseparable region as follows. \\\vspace{-8pt}\\
 \indent\textbf{Definition 1:} \emph{Suppose $\Theta$ is the searching space of the parameter $\theta$, the dissimilar inseparable region $\mathcal{N}^a_i$ and similar inseparable region $\mathcal{N}^b_i$ of the query sample $\textbf{x}_i$ are defined as follows.}
 \begin{equation}\label{inseq_reg}
 \small
 \mathcal{N}^a_i=  \bigcap_{\theta \in \Theta}\mathcal{A}_{i}(\theta), \quad\quad \mathcal{N}^b_i=  \bigcap_{\theta \in \Theta}\overline{\mathcal{B}_{i}(\theta)}
\end{equation}
\emph{where $\overline{\mathcal{B}_{i}(\theta)} = \mathbb{R}^d - \mathcal{B}_{i}(\theta)$}.\\\vspace{-7pt}\\
\indent According to Definition 1, a dissimilar sample $\textbf{x}_l \in \mathcal{D}_i$ located in $\mathcal{N}^a_i$ (or a similar sample $\textbf{x}_j \in \mathcal{S}_i$ located in $\mathcal{N}_i^b$) could not be excluded by $\mathcal{A}_i(\theta)$(or contained by $\mathcal{B}_i(\theta)$) by searching $\theta$. Thus, the dissimilar sample $\textbf{x}_l$ (or similar sample $\textbf{x}_j$) can not be separated from the similar samples (or dissimilar samples). The more dissimilar (or similar) samples located in $\mathcal{N}^a_i$ (or $\mathcal{N}^b_i$) are, the more serious the inseparable problem of metric learning is.\\
\indent Since $f_\theta(\textbf{x})$ can be decomposed as $f_\theta(\textbf{x}) = \textbf{L}^Tf'_\theta(\textbf{x})$ where $f'_\theta(\textbf{x})=(\textbf{L}^{-1}f_\theta(\textbf{x}))$, a distance metric learning can be divided into two procedures: firstly projecting the training set by $f'_\theta(\textbf{x})$, then performing linear distance metric learning in the projection space of $f'_\theta(\textbf{x})$. Thus, exploring $\mathcal{N}^a_i$ and $\mathcal{N}^b_i$ of linear model can help us to understand the inseparable problem of metric learning, and \textbf{Theorem 1} is introduced as follows.\\\vspace{-5pt}\\
\indent \textbf{Theorem 1:}\emph{ If the projection function $f_\theta(\textbf{x}) = \textbf{L}^T\textbf{x}$, the dissimilar and similar inseparable region of $\textbf{x}_i$ are $\mathcal{N}^a_i = \{\textbf{x}| \textbf{x}=\sum_{j=1}^{\vert\mathcal{S}_i\vert} r_j (\textbf{x}_j-\textbf{x}_i) + \textbf{x}_i,\sum_{j=1}^{\vert\mathcal{S}_i\vert} \vert r_j\vert<1 \}$, and $\mathcal{N}^b_i = \{\textbf{x}| \textbf{x}=\sum_{l=1}^{\vert\mathcal{D}_i\vert} r_l (\textbf{x}_l-\textbf{x}_i) + \textbf{x}_i,\sum_{l=1}^{\vert\mathcal{D}_i\vert} \vert r_l\vert>1 \}$.}\\\vspace{-5pt}\\
\indent The proof is presented in the supplemental material. Let us take $\mathcal{N}^a_i$ as an example. According to\textbf{ Theorem 1}, when $\vert\mathcal{S}_i\vert = 1,2,3$, $\mathcal{N}^a_i$ is line segment, parallelogram, and polyhedron, respectively. The graphically illustration is presented in Fig.\ref{fig_neb_r_line}. Then, we can find the following observations from \textbf{Theorem 1}.
 \begin{itemize}
  \item Due to $\sum_{j\in \mathcal{S}_i} \vert r_j\vert<1$, $\mathcal{N}^a_i$ is an expansion of the region whose vertexes are the samples in $\mathcal{S}_i$. As shown in Fig.\ref{fig_neb_r_line}, the line segment $\textbf{x}_i\textbf{x}_j$ expands to the line segment $\textbf{x}_j\hat{\textbf{x}_j}$, the triangle $\textbf{x}_i\textbf{x}_j^1\textbf{x}_j^2$ expands to the parallelogram $\hat{\textbf{x}}_j^1\hat{\textbf{x}}_j^2\textbf{x}_j^1\textbf{x}_j^2$, and so on. The region expansion makes that even if two classes are not intersected with each other, there would be still many dissimilar samples located within $\mathcal{N}^a_i$ to become inseparable samples. This is why distance metric learning is easy to encounter inseparable problem than the margin-based classification model.
       \item Since larger area of $\mathcal{N}^a_i$ (or $\mathcal{N}^b_i$) means more dissimilar (or similar) samples would locate within it, thus more inseparable samples would be produced with the area of $\mathcal{N}^a_i$ (or $\mathcal{N}^b_i$) increasing. As a result, the severity of inseparable problem of metric learning relies on two factors which determine the area of $\mathcal{N}^a_i$ (or $\mathcal{N}^b_i$): the number of samples in $\mathcal{S}_i$ (or $\mathcal{D}_i$) determining the dimension of $\mathcal{N}^a_i$ (or $\mathcal{N}^b_i$), and the distance $\vert \textbf{x}_i-\textbf{x}_j\vert$ measuring the divergence of samples in $\mathcal{S}_i$ centered at $\textbf{x}_i$. Thus, when the training samples are more, and dispute with larger divergence, the inseparable problem would be more serious.
 \end{itemize}
\indent \vspace{-10pt}\\
\indent Besides the inseparable regions, that using the same parameter $\theta$ to depict the discriminant information of different query samples would also produce many inseparable triplets. Suppose the feasible solution set of triplet $(\textbf{x}_i,\textbf{x}_j,\textbf{x}_l)$ is $\Theta_{ijl} = \{\theta| d_\theta(\textbf{x}_i,\textbf{x}_j)<d_\theta(\textbf{x}_i,\textbf{x}_l)\}$. Thus, $\Theta_{i} =\bigcap_{j\in \mathcal{S}_i,l\in \mathcal{D}_i}\Theta_{ijl}$ is the feasible set of query $\textbf{x}_i$ to separate its similar and dissimilar samples. For the whole training set, the feasible set is $\Theta =\bigcap_{i=1}^N\Theta_i$. Since query samples from different class may have different feasible sets, the successive intersection operations on those different feasible set, would make the global feasible set $\Theta$ be empty. This indicates that some of dissimilar samples would become inseparable, even if they are separable for every single query. This type of inseparable problem can also be explained geometrically. Take the samples presented in Fig.\ref{fig_neb_r_line}(d) as example, the inseparable region of $\textbf{x}_2$ is $\hat{\mathcal{N}}^a_2 = \bigcap_\theta\{\textbf{x}|d_\theta(\textbf{x}_2,\textbf{x}_1)>d_\theta(\textbf{x}_2,\textbf{x}_4)\bigwedge d_\theta(\textbf{x},\textbf{x}_4)<d_\theta(\textbf{x},\textbf{x}_2)\}$. This is another reason why the metric learning easily suffers from serious inseparable problem.\\
\indent As discussed above, a simple training dataset may produce lots of inseparable samples. In the supplemental material, we show a separate data set with $6$ samples under margin-based classification producing $16.7\%\thicksim 33.4\%$ inseparable samples under triplet-based criterion. Since the dimension and enlargement of inseparable regions are all increased with the amount of training samples, it is difficult to pull all of the inseparable samples out of inseparable regions by the non-linear projection trained by the training samples themselves. Thus, the non-linear metric learning could also suffer from serious inseparable problem.
{
\subsection{Theoretical Analysis on the Inseparable Problem of Metric learning}
\label{taipmel_ss}
\indent{In this subsection, we analyse the inseparable problem of metric learning from the perspective of Lipschitz constant which is a widely-used measurement to evaluate how hard the training of a non-linear learning model is \cite{kuhn2014nonlinear,virmaux2018lipschitz,arjovsky2017wasserstein}. Before doing this, we introduce several definitions as follows.}\\\vspace{-5pt}\\
\indent \textbf{Definition 2:}\cite{dong2019learning} \emph{Let $(\mathcal{X},d^{\mathcal{X}})$ and $(\mathcal{Y},d^{\mathcal{Y}})$ be two metric spaces, the \textbf{Lipschitz constant} of a function $f$ is defined as:}
\begin{equation}\label{eq_Lpc}
\small
Lip(f)= \max_{\textbf{x}_i,\textbf{x}_j\in \mathcal{X}:i\neq j}\frac{d^{\mathcal{Y}}(f(\textbf{x}_i),f(\textbf{x}_j))}{d^{\mathcal{X}}(\textbf{x}_i,\textbf{x}_j)}
\end{equation}
\emph{where $d^{\mathcal{X}}$ and $d^{\mathcal{Y}}$ are the distance metrics on sets $\mathcal{X}$ and $\mathcal{Y}$, respectively.}
\\\vspace{-8pt}\\
\indent \textbf{Definition 3:} \emph{Given $C$ classes of samples $\{\mathcal{X}_c\}_{c=1}^C$, the \textbf{class-gap} between the $p$-th class $\mathcal{X}_p$ and the $p$-th class  $\mathcal{X}_q$ is defined as $\delta_{pq}=\min_{\textbf{x}_i\in \mathcal{X}_p,\textbf{x}_j \in \mathcal{X}_q}d(\textbf{x}_i,\textbf{x}_j)$, where $d(\textbf{x}_i,\textbf{x}_j)$ is the distance between $\textbf{x}_i$ and $\textbf{x}_j$. The class-gap of the whole data set is defined as $\delta = \min_{p\neq q}{\delta_{pq}}$}.\\\vspace{-5pt}\\
\indent \textbf{Theorem 2:} \emph{Suppose the class-gap of data set $\{\mathcal{X}_c\}_{c=1}^C$ is $\delta$, after projecting by the function $f(\textbf{x})$, the class-gap of the data set becomes $\hat{\delta}$. Thus, the Lipschitz constant of $f({\textbf{x}})$ satisfies the condition as follows:}\vspace{-5pt}
\begin{equation}\label{the_eq2}
\small
Lip(f)>\frac{\hat{\delta}}{\delta}
\end{equation}
\vspace{-2pt}
\indent \emph{Proof}:{ According to the \textbf{Definition 3}, we can find two samples $\textbf{x}_i$ and $\textbf{x}_l$ to let $d({\textbf{x}}_i,{\textbf{x}}_l) = \delta$. According to the definition of Lipschitz constant, we have $Lip(f) > \frac{d(f({\textbf{x}}_i),f({\textbf{x}}_l))}{d({\textbf{x}}_i,{\textbf{x}}_l)}=\frac{d(f({\textbf{x}}_i),f({\textbf{x}}_l))}{\delta}$. Similarly, according to the Definition 3, there is $d(f(\textbf{x}_i),f(\textbf{x}_l)) > \hat{\delta}$. Thus, we obtain $Lip(f)>\frac{\hat{\delta}}{\delta}$.
$\Box$ \\\vspace{-5pt}\\}
\indent \textbf{{Proposition 1}}: \emph{If the class-gap of a data set in the original space is very small, i.e., $\delta \rightarrow 0$, the Lipschitz constant of the desired projection $f_\theta(\textbf{x})$ obtained by triplet constraints satisfies the condition $Lip(f_\theta) > \frac{\hat{\delta} }{\delta } \rightarrow +\infty$.} \\\vspace{-5pt}\\
{\indent \emph{Proof:} One of the purposes of metric learning is to increase the inter-class variance. So after metric learning, the class-gap of a data set should be enlarged, and there is $\hat{\delta}\gg 0$. According to \textbf{Theorem 2}, when $\delta\rightarrow 0$, there is $Lip(f_\theta) > \frac{\hat{\delta} }{\delta } \rightarrow +\infty$.
 $\Box$\\\vspace{-10pt}\\}
{\indent For linear metric learning, considering $f_{\theta}(\textbf{x})=\frac{1}{s_m}\textbf{L}^T\textbf{x}$ where $s_m$ is the largest single value of $\textbf{L}$, the Lipschitz constant of $f_{\theta}(\textbf{x})$ satisfies the condition $Lip(f_\theta) < \frac{1}{s_m}\sqrt{tr(\textbf{L}^T\textbf{L})}\leq d$, where $d$ is the rank of $\textbf{L}$ \cite{dong2019learning}. Since $\frac{1}{s_m}$ does not change the learning ability of $f_{\theta}(\textbf{x})$, the Lipschitz constant of linear metric learning has a very small upper-bound. According to \textbf{{Proposition 2}}, the triplet constraints may lead to $Lip(f_L) \rightarrow +\infty$ when the class-gap is small. In this case, the learning ability of the linear projection with $Lip(f_\theta)<d$ is not enough. Consequently, the linear metric learning would suffer from inseparable problem.\\
\indent In deep metric learning, the training images from different classes are often mixed with each other, which means the class-gap before projecting approximates $0$, i.e., $\delta \rightarrow 0$. Thus, according to \textbf{{Proposition 2}}, the $Lip(f_\theta) \rightarrow +\infty$. However, literatures \cite{sun2019optimization,kuhn2014nonlinear} reveal that the convergence of a non-linear optimization problem solved by gradient descent methods requires a upper-bounded Lipschitz constant of the desired projection. This may conflict to the requirement of deep metric learning, i.e, $Lip(f_\theta) \rightarrow +\infty$. In this situation, the desired deep neural network could not be obtained by gradient descent methods. Thus, we claim that the deep metric learning also suffers from inseparable problem.\\
\indent Thus, the role of the informative sample mining is to remove the far similar samples and the near dissimilar samples of each anchor, which would enlarge the class-gap $\delta$ in the original space or to reduce the class-gap $\hat{\delta}$ in the projected space to reduce the Lipschitz constant of the desired projection. In this way, the training of the model would become more easy.}
\begin{figure}[t]
 \centering
  \includegraphics[width=0.7\linewidth]{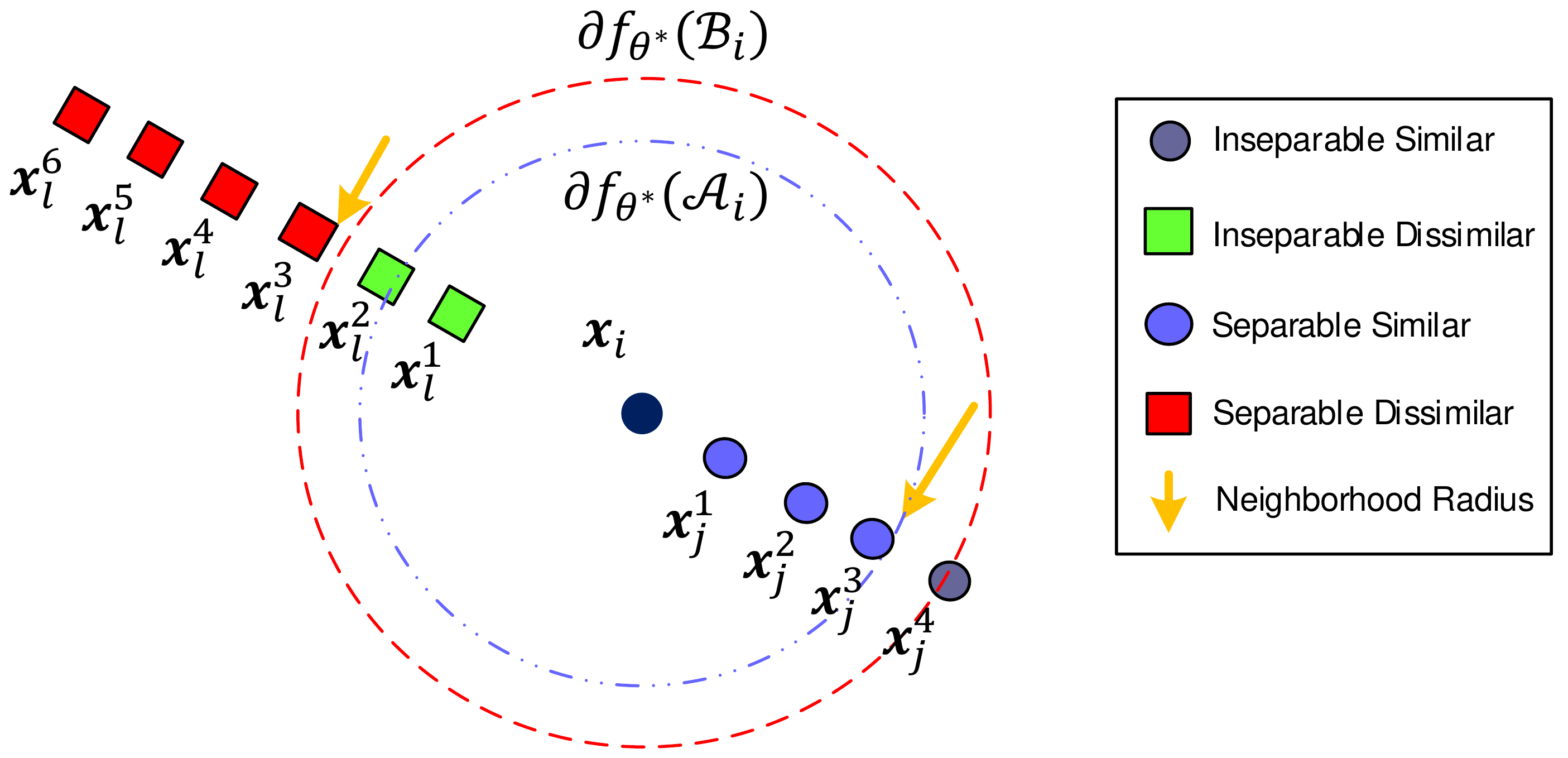}
\vspace{-9pt}\caption{{The illustration of the inseparable and separable samples of the query $\textbf{x}_i$ in the projected space. $\partial f_{\theta^*}(\mathcal{A}_i)$ and $\partial f_{\theta^*}(\mathcal{B}_i)$ are the boundaries of the neighborhoods $\mathcal{A}_i$ and $\mathcal{B}_i$ in projected space, respectively.}}
  \label{fig_neb1}
\end{figure}
\vspace{-3pt}\section{Adaptive Neighborhood Metric Learning}
\subsection{Motivation}
\label{andml_mtv}
\indent {{Due to the close dependence on the learned projection function, the inseparable samples can be detected in the projected space. Suppose $\theta^*$ is the optimal parameter of the projection function $f_{\theta^*}(\textbf{x})$, the projections of the neighborhoods $\mathcal{A}_i$ and $\mathcal{B}_i$ of the anchor $\textbf{x}_i$ are $f_{\theta^*}(\mathcal{A}_i)$ and $f_{\theta^*}(\mathcal{B}_i)$, respectively. According to the definitions of $\mathcal{A}_i$ and $\mathcal{B}_i$, only and all the similar samples are located in them after projecting. Thus, the similar samples located out of $f_{\theta^*}(\mathcal{A}_i)$ and the dissimilar samples located in $f_{\theta^*}(\mathcal{B}_i)$ are inseparable samples. The illustration of the inseparable samples in the projected space is shown in Fig. \ref{fig_neb1}. \\
 \indent As seen in Fig. \ref{fig_neb1}, since $f_{\theta^*}(\textbf{x}_l^1)$ and $f_{\theta^*}(\textbf{x}_l^2)$ are located in $f_{\theta^*}(\mathcal{B}_i)$, and $f_{\theta^*}(\textbf{x}_j^4)$ is located out of $f_{\theta^*}(\mathcal{A}_i)$, they are the inseparable samples. Obviously, $\textbf{x}_l^2$ is the farthest inseparable dissimilar sample from the query $\textbf{x}_i$, thus all the dissimilar samples with smaller distance than $d_{\theta^*}(\textbf{x}_i,\textbf{x}_l^2)$ from $\textbf{x}_i$ are the inseparable dissimilar samples. Similarly, $\textbf{x}_j^4$ is the nearest inseparable similar sample from the query $\textbf{x}_i$, thus, all the similar samples with larger distance than $d_{\theta^*}(\textbf{x}_i,\textbf{x}_j^4)$ are inseparable similar samples. This motivates us that, for the query $\textbf{x}_i$, we could find two integers $K_s$ and $K_d$ a priori to let the $K_s$ farthest similar samples and the $K_d$ nearest dissimilar samples are the inseparable samples.}}\\
 \indent This procedure is much similar to the informative sample mining, especially the semi-hard sample mining. Thus, we argue that one purpose of informative sample mining is to remove the inseparable samples. As the benefit, why Triplet loss can be improved by semi-hard sample mining instead of the hard sample mining can be well explained. However, the inseparable samples are often mixed with the hard samples which produce gradient with large magnitudes. This makes informative sample mining often come into being some side-effects. For example the semi-hard sample mining will lead to the instability of objective function since it removes the hardest samples, while the hard sample mining preserves the most inseparable samples since the hardest samples are likely inseparable. Those side-effects would hurt the performance of metric learning.
\begin{figure}[t]
 \centering
  \includegraphics[width=0.7\linewidth]{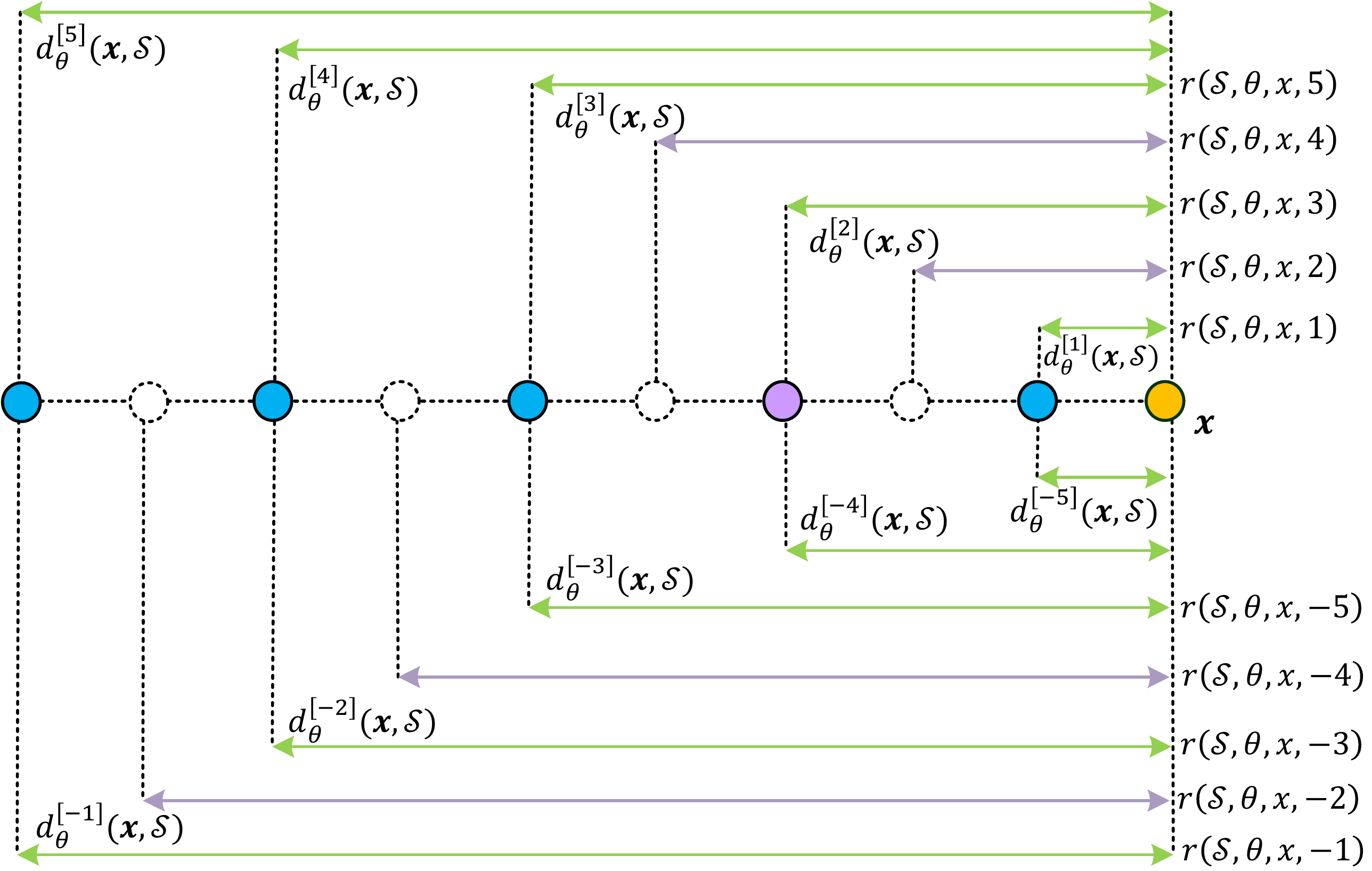}
\vspace{-9pt}\caption{The graphical illustration of values of the function $r(\mathcal{S},\theta,\textbf{x}, K\alpha)$ presented in Eq.(\ref{neighbor_di}). For better illustration, all the samples are set on a line, and the distance set $\{d_{\theta}(\textbf{x},\textbf{x}_j)|\textbf{x}_j\in \mathcal{S}\}$ forms a arithmetic sequence. The blue circles are the samples in the set $\mathcal{S}$. The white circles are the positions estimated by $r(\mathcal{S},\theta,\textbf{x}, K\alpha)$.
}
\label{fig_nebrsxv}
\end{figure}
 \vspace{-10pt}
 \subsection{Formulation}\label{section_42}
\indent Let symbols $d_\theta^{[k]}(\textbf{x},\mathcal{S})$ and $d_\theta^{[-k]}(\textbf{x},\mathcal{S})$ denote the $k$-th smallest value and the $k$-th largest value in $\{d_\theta(\textbf{x},\textbf{x}_j)| \textbf{x}_j \in \mathcal{S}\}$, respectively. As the discussion in subsection \ref{andml_mtv}, there exist two integers $K_s$ and $K_d$ to let $d_{\theta^*}^{[K_s]}(\textbf{x}_i,\mathcal{S}_i)$ and $d_{\theta^*}^{[K_d]}(\textbf{x}_i,\mathcal{D}_i)$ be thresholds to identify the separable similar and dissimilar samples, respectively. Since those thresholds determined by a single sample are less robust to noises, we design more robust ones by using the following equation:
\begin{equation}\label{eq1_rad}
\small
\begin{split}
r(\mathcal{S},\theta,\textbf{x}, K\alpha) &=  \frac{1}{K}\sum_{k=1}^{K}d_\theta^{[k\alpha]}(\textbf{x},\mathcal{S})\\
\end{split}
\end{equation}
 \emph{where $\mathcal{S}$ is a sample set, $\alpha \in \{-1,1\}$. As depicted in Fig.\ref{fig_nebrsxv}, the range of $r(\mathcal{S},\theta,\textbf{x},K\alpha)$ is $ [d^{[1]}_{\theta}(\textbf{x},\mathcal{S})),d^{[-1]}_{\theta}(\textbf{x},\mathcal{S})]$ with $K\alpha \in \{\pm1,\cdots,\pm \vert \mathcal{S}\vert\}$. Therefore, there are two couples $(K_1,\alpha_1)$ and $(K_2,\alpha_2)$ to let terms $r(\mathcal{S}_i,\theta^*,\textbf{x}_i, K_1\alpha_1)$ and $r(\mathcal{D}_i,\theta^*,\textbf{x}_i, K_2\alpha_2)$ approximate the thresholds $d_{\theta^*}^{[K_s]}(\textbf{x}_i,\mathcal{S}_i)$ and $d_{\theta^*}^{[K_d]}(\textbf{x}_i,\mathcal{D}_i)$ to identify the separable samples.}\\
\vspace{-0pt}\indent Besides finding the separable samples, $\theta^*$ also achieves the goal of metric learning, i.e., separating the separable dissimilar and similar samples by the boundary of the neighborhood of each $\textbf{x}_i$. Thus, we can redefine two types of neighborhoods of query sample $\textbf{x}_i$ as follows:
\begin{equation}\label{neighbor_si}
\small
\mathcal{A}_i(\theta^*) = \{\textbf{x}|d_{\theta^*}(\textbf{x}_i,\textbf{x}) \leq r(\mathcal{S}_i,\theta^*,\textbf{x}_i, K_1\alpha_1)\}
\end{equation}
\vspace{-9pt}
 \begin{equation}\label{neighbor_di}
 \small
\quad\mathcal{B}_i(\theta^*) = \{\textbf{x}|d_{\theta^*}(\textbf{x}_i,\textbf{x}) \leq r(\mathcal{D}_i,\theta^*,\textbf{x}_i, K_2\alpha_2)\}
\end{equation}
The goal of metric learning can be depicted mathematically as follows:
\begin{equation}\label{eqd}
\small
r(\mathcal{S}_i,\theta^*,\textbf{x},K_1\alpha_1) \leq r(\mathcal{D}_i,\theta^*,\textbf{x},K_2\alpha_2)
\end{equation}
The geometry illustration of this condition is shown in Fig. \ref{fig_neb1}.\\
\indent Obviously, when $K_1\alpha_1$ and $K_2\alpha_2$ are specialized previously, Eq. (\ref{eqd}) can be used to construct a metric learning formulation since it constrains the optimal solution $\theta^*$. In fact, the term $r(\mathcal{S},\theta,\textbf{x},K\alpha)$ in Eq. (\ref{eq1_rad}) can be rewritten in the optimization form as follows:
\begin{equation}\label{eq1_rad1}
\small
\begin{split}
r(\mathcal{S},\theta,\textbf{x}, K\alpha) &= g(\alpha)\min_{\mathcal{K} \in \Pi^{K}(\mathcal{S})}\frac{1}{\vert \mathcal{K}\vert}\sum_{j\in \mathcal{K}}d_\theta(\textbf{x},\textbf{x}_j)+ \quad\quad\\
  &\quad g(-\alpha)\max_{\mathcal{K} \in \Pi^{K}(\mathcal{S})}\frac{1}{\vert \mathcal{K}\vert}\sum_{j\in \mathcal{K}}d_\theta(\textbf{x},\textbf{x}_j)\\
\end{split}
\end{equation}
 where $\mathcal{K}$ is a subset of $\mathcal{S}$ with $\vert \mathcal{K}\vert = K$ and $\Pi^{K}(\mathcal{S}) = \{\mathcal{K}|\mathcal{K} \subset \mathcal{S}, \vert\mathcal{K}\vert=K\}$, $\vert\mathcal{K}\vert$ is the number of elements in $\mathcal{K}$. $g(\alpha)$ ($\alpha \in \{1,-1\}$) is an indicator function with $g(1) = 1$ and $g(-1) = 0$.\\
\indent Rewriting the inequality in Eq.(\ref{eqd}) by using the Eq.(\ref{eq1_rad1}), the formulation of adaptive neighborhood metric learning is obtained as follows:
   \begin{equation}\label{imtco2}
   \small
   \setlength{\jot}{-0pt} % affecting the line spacing in the environment
  \begin{split}
  & \min_{\theta} \sum_{i=1}^N \ell\left(\min_{\mathcal{K}^d\in \Pi^{K_2}(\mathcal{D}_i)}\frac{1}{\vert \mathcal{K}^{d}\vert}\sum_{l\in \mathcal{K}^{d}}d_\theta(\textbf{x}_i,\textbf{x}_l)\right. \\
  &\quad\quad\quad\quad\quad\left.- g(\alpha)\min_{\mathcal{K}^a \in \Pi^{K_1}(\mathcal{S}_i)}\frac{1}{\vert \mathcal{K}^a\vert}\sum_{j\in \mathcal{K}^s}d_\theta(\textbf{x}_i,\textbf{x}_j)\right.\\
  &\quad\quad\quad\quad\left. -  g(-\alpha)\max_{\mathcal{K}^s \in \Pi^{K_1}(\mathcal{S}_i)}\frac{1}{\vert \mathcal{K}^s\vert}\sum_{j\in \mathcal{K}^s}d_\theta(\textbf{x}_i,\textbf{x}_j)\right)
  \end{split}
 \end{equation}
 where $\ell\left(x\right)$ is a loss function penalizing large $x$, $\mathcal{K}^s$ is a subset of $\mathcal{S}_i$ consisting of $K_1$ samples while $\mathcal{K}^d$ is a subset of $\mathcal{D}_i$ consisting of $K_2$ samples. For the dissimilar set $\mathcal{D}_i$, we set $\alpha = 1$ because the inseparable dissimilar sample should be less than the separable similar samples, or the metric learning is meaningless.\\
 \indent As seen from the above formulation, with the fixed parameters $(K_1,\alpha_1)$ and $(K_2,\alpha_2)$, the proposed method automatically selects the \emph{separable samples} in $\mathcal{S}_i$ and $\mathcal{D}_i$ to support the neighborhoods $\mathcal{A}_i$ and $\mathcal{B}_i$. Meanwhile, the optimal metric parameter $\theta^*$ is solved. That is why we call our method as \emph{\textbf{Adaptive Neighborhood Metric Learning} (ANML)}. Thus, in our method, the informative sample mining strategies can only focus on removing the samples producing gradients with magnitude close to zero.\\
  \indent However, the proposed formulation in Eq.(\ref{imtco2}) can not be solved by the gradient descent method since its objective function is non-continuous.
  \vspace{-10pt} \subsection{Continuous Proxy of ANML}
In this section, we propose the continuous proxy of ANML. Before doing this, we introduce two useful lemmas as follows.\\\vspace{-9pt}\\
\indent \textbf{Lemma 1.} \emph{Given a series of numbers $\left\{a_i\right\}_{i=1}^n$, without loss of generality, they are listed in ascending order, i.e., $ a_1\leq a_2\leq\cdots\leq a_n$. Considering the \textbf{log-exp mean function} presented as follows:}
\vspace{-4pt}
 \begin{equation}\label{lm1}
 \small
 \begin{split}
&b\left(\gamma\right) = -\frac{1}{\gamma}log\left(\frac{\sum_{i=1}^ne^{-\gamma a_i}}{n}\right),\\
 \end{split}
 \end{equation}
\emph{there exist the following relationships:}
 \begin{equation}\label{lm11}
 \begin{split}
 \small
 \lim_{\gamma\rightarrow 0}\!b\left(\gamma\right)=\sum_{i=1}^n\frac{a_i}{n},\lim_{\gamma\rightarrow+\infty}\!b\left(\gamma\right)=a_1,\lim_{\gamma\rightarrow-\infty}\!b\left(\gamma\right)=a_n
 \end{split}
 \end{equation}
 \indent \textbf{Lemma 2.} \emph{The $b(\gamma)$ defined in Eq.(\ref{lm1}) is a monotone decreasing function with respective to $\gamma$, i.e., for $\gamma_1<\gamma_2$, there is $b(\gamma_1)<b(\gamma_2)$.}\\\vspace{-5pt}\\
 \indent The proofs of {\textbf{Lemma 1}}, {\textbf{Lemma 2}} are presented in the supplemental materials. After giving those two lemmas, we can obtain the\emph{ \textbf{Theorem 3} }as follows.\\\vspace{-5pt}\\
 \indent \textbf{Theorem 3.} \emph{Given a set of numbers listed in ascending order, i.e., $a_1<a_2<\cdots<a_n$ and an integer $K \leq n$, there exist one and only one $\gamma_1^*>0$ and $\gamma_2^*<0$ to let $b(\gamma)$ defined in Eq.(\ref{lm1}) hold the following equations, respectively.}
 \begin{equation}\label{lemma2}
 \small
 \begin{split}
b(\gamma_1^*)=\frac{1}{K}\sum_{k=1}^Ka_k, \quad b(\gamma_2^*)=\frac{1}{K}\sum_{k=1}^Ka_{(n-k+1)}
 \end{split}
 \end{equation}
\indent The proof of {\textbf{Theorem 3}} is also presented in the supplemental materials. The correspondence relationship between the $K$ and $b(\gamma)$ in the Eq.(\ref{lemma2}) is shown in Fig. \ref{fig_b_fun}. {\textbf{Theorem 3}} implies that the continuous\emph{ log-exp mean function} $b(\gamma)$ can be used to estimate the term defined in Eq.(\ref{eq1_rad}).\\
 \indent Let us define two functions $d^s_{x_i}(\theta,\gamma)$ and $d^d_{x_i}(\theta,\gamma)$ as follows:
\begin{equation}\label{intro_func}
\small
\begin{split}
d^s_{x_i}(\theta,\gamma)&= \frac{1}{-\gamma}\log\left(\frac{1}{\vert\mathcal{S}_i\vert}\sum_{j\in\mathcal{S}_i}{e^{-\gamma d_\theta(\textbf{x}_j,\textbf{x}_i)}}\right)\quad\\
\end{split}
\end{equation}
\vspace{-4pt}
\begin{equation}
\small
\label{intro_func22}
\begin{split}
\quad\quad d^d_{x_i}(\theta,\gamma)&= \frac{1}{-\gamma}\log\left(\frac{1}{\vert\mathcal{D}_i\vert}\sum_{l\in\mathcal{D}_i}{e^{-\gamma d_\theta(\textbf{x}_l,\textbf{x}_i)}}\right)\quad\\
\end{split}
\end{equation}
where $\vert \mathcal{S}_i\vert$ and $\vert \mathcal{D}_i\vert$ are the numbers of samples in $\mathcal{S}_i$ and $\mathcal{D}_i$, respectively.
 \begin{figure}[t]
 \centering
  \includegraphics[width=0.65\linewidth]{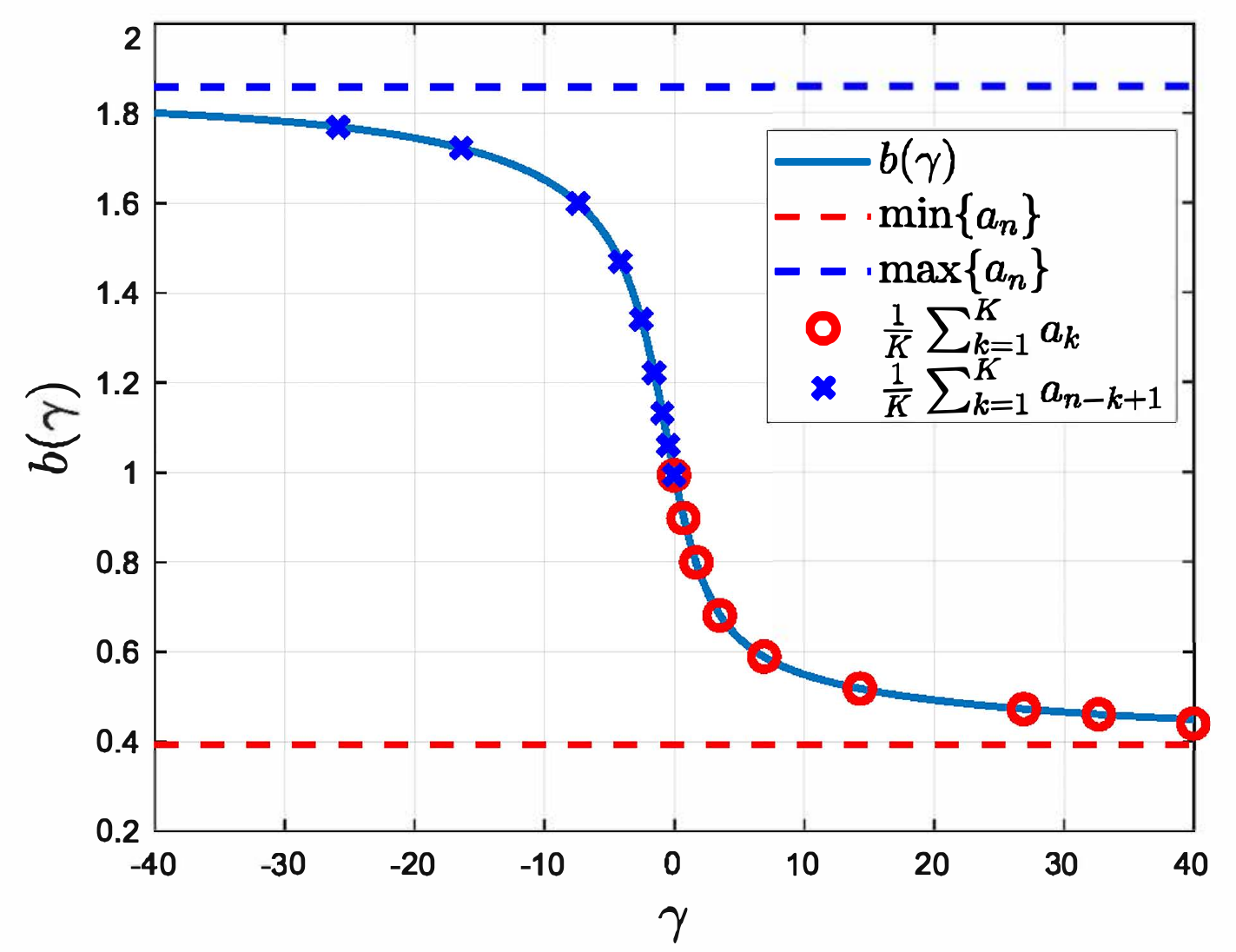}
\vspace{-10pt} \caption{\emph{The illustration of the correspondence relationship between the $\frac{1}{K}\sum_{k=1}^Ka_{k}$ (or $\frac{1}{K}\sum_{k=1}^Ka_{n-k+1}$) and $b(\gamma)$. The number series $\{a_i\}_{i=1}^n$ is generalized randomly with $n=10$ and ranked with ascending order. From left to right, 'x' dots represent the values of $\frac{1}{K}\sum_{k=1}^Ka_{n-k+1}$ with $K=2,3,\cdots,10$, respectively. Similarly, 'o' dots represent the values of $\frac{1}{K}\sum_{k=1}^Ka_{k}$ with $K=10,9,\cdots,2$, respectively.}}
  \label{fig_b_fun}
\end{figure}
According to \emph{\textbf{Theorem 3}}, for data sets $\mathcal{S}_i$ and $\mathcal{D}_i$, there are two values $\gamma_1 \in \mathbb{R}$ and $\gamma_2 \in \mathbb{R}$ to make the following equations be established.
\begin{equation}
\label{into_pqeq}
 \begin{split}
 d^s_{x_i}(\theta,\gamma_1) &= r(\mathcal{S}_i,\theta,\textbf{x}_i, K_1\alpha_1)\\
  d^d_{x_i}(\theta,\gamma_2) &= r(\mathcal{D}_i,\theta,\textbf{x}_i, K_2\alpha_2)
 \end{split}
 \end{equation}
\indent By substituting Eq.(\ref{into_pqeq}) into Eq.(\ref{eqd}), a new constraint to describe the separation of separable similar and dissimilar samples of the $i$-th query sample is presented as follows:
\begin{equation}\label{constaint2}
\small
\begin{split}
&\frac{1}{-\gamma_1}\log\left(\frac{1}{\vert\mathcal{S}_i\vert}\sum_{j\in\mathcal{S}_i}{e^{-\gamma_1d_\theta(\textbf{x}_i,\textbf{x}_j)}}\right)< \\ &\quad\quad\quad\quad\quad\quad\quad\frac{1}{-\gamma_2}\log\left(\frac{1}{\vert\mathcal{D}_i\vert}\sum_{l\in\mathcal{D}_i}{e^{-\gamma_2 d_\theta(\textbf{x}_i,\textbf{x}_l)}}\right)
\end{split}
\end{equation}
\indent By replacing the constraint presented in Eq.(\ref{eqd}) with the one in Eq.(\ref{constaint2}), the non-continuous optimization problem described in Eq.(\ref{imtco2}) can be transformed into a continuous one. Similar to that Triplet loss can be used into both linear and non-linear models, Eq.(\ref{constaint2}) can also be applied to both of the tasks by selecting different similarity functions to replace $d_\theta(\textbf{x}_i,\textbf{x}_j)$. \\
%\indent The implementations of ANML of different similarity functions may require different optimization strategies. For example, when $d_\theta(\textbf{x}_i,\textbf{x}_j)$  is a linear function with respected to $\theta$, ANML can be trained directly by gradient descent methods. When $d_\theta(\textbf{x}_i,\textbf{x}_j)$ is the similarity function associated with deep features, the model is often solved by stochastic gradient descent methods, in which the training data should be split randomly into several groups called as mini-batches. Thus, for different optimization strategies, the objective functions of ANML should be a little different. In the following sections, we would give the implementations of ANML for learning Mahalanobis distance metric and deep feature embedding, respectively.
\vspace{-5pt}\section{Learning Mahalanobis Distance Metric}
\vspace{-3pt}\subsection{Formulation}
%\indent The linear model can help us to find some useful properties of the loss function of a metric learning algorithm, such as whether the loss function is convex or not.
\indent By replacing the distance function $d_\theta(\textbf{x}_i,\textbf{x}_j)$ with the squared distance $d_{\textbf{M}}(\textbf{x}_i,\textbf{x}_j) = (\textbf{x}_i -\textbf{x}_j)^T\textbf{M}(\textbf{x}_i-\textbf{x}_j)$ where $\textbf{M} \in \mathbb{R}^{d\times d}$, the formulation of the linear adaptive neighborhood metric learning (LANML) is presented as follows:
   \begin{equation}\label{optim_continous2}
 \begin{split}
  \min_{\textbf{M}\succeq 0} \sum_{i=1}^N & \ell\left(-\frac{1}{\gamma_1} \log[\frac{1}{\vert\mathcal{S}_i\vert}\sum_{j\in\mathcal{S}_i}{e^{-\gamma_1 d_{{\textbf{M}}}(\textbf{x}_i,\textbf{x}_j)}}]+\right.\\
  &\quad\left. \frac{1}{\gamma_2} \log[\frac{1}{\vert\mathcal{D}_i\vert}\sum_{l\in\mathcal{D}_i}{e^{-\gamma_2d_{{\textbf{M}}}(\textbf{x}_i,\textbf{x}_l)}}]\right)+\lambda\Omega(\textbf{M})
 \end{split}
 \end{equation}
where $\lambda$ is the parameter of the regularization term $\Omega(\textbf{M})$. Here, we set $\gamma_2>0$, since the number of inseparable dissimilar samples is supposed less than that of separable similar samples.\\
\indent To demonstrate whether LANML is convexity or not, the following {\emph{Lemma 3}} is introduced, which also shows the superiority of using the \emph{log-exp mean function} to construct the metric learning algorithm.\\\vspace{-5pt}\\
\indent \textbf{Lemma 3}: \emph{When} $\gamma_1 < 0$ and $\gamma_2 > 0$, \emph{the optimization problem presented in} Eq.(\ref{optim_continous2}) \emph{is a convex optimization problem.}\\\vspace{-5pt}\\
 \indent The proof is appended in the supplemental materials. \textbf{Lemma 3} explains why we choose the function $b(\gamma)$ in Eq.(\ref{lm1}) to construct the continuous proxy of ANML in Eq.(\ref{imtco2}). Actually, besides $b(\gamma)$, there exists other function to let the objective of Eq.(\ref{imtco2}) be continuous, e.g., the function $\hat{b}(\gamma) = (\frac{1}{N}\sum_{i=1}^N (a_i)^\gamma)^{\frac{1}{\gamma}}$ has the same property as the function $b(\gamma)$ stated in \textbf{Theorem 3}. However, $\hat{b}(\gamma)$ can not make the formulation presented in Eq.(\ref{imtco2}) be convex. That is the reason why we propose \emph{log-exp mean function} to reformulate ANML.\\
\vspace{-10pt}\subsection{Relationship between LANML and Other Linear Models}
\label{relationship_ann_other}
\indent In this section, we will discuss the relationships between the proposed LANML and other famous linear metric learning methods including large margin nearest neighbor (LMNN) and neighborhood component analysis (NCA), respectively.\\
\vspace{-10pt}\subsubsection{The Connection between LANML and LMNN}
 \textbf{Proposition 3:} \emph{When $\gamma_1\rightarrow -\infty$, $\gamma_2\rightarrow +\infty$ and the similarity set $\mathcal{S}_i$ is selected as the target neighbors, the model described in Eq.(\ref{optim_continous2}) is a convex improvement of LMNN.}\\\vspace{-5pt}\\
 \emph{Proof:} \indent {When} $\gamma_1 \rightarrow -\infty$,\emph{ the function} $d^s_{{x}_i}(\textbf{M},\gamma_1)= \sup\{d_\textbf{M}(\textbf{x}_i,\textbf{x}_j)| \textbf{x}_j \in \mathcal{S}_i\} = d^{max}_i$. In this case, the objective of LANML is to let most of the samples in $\mathcal{D}_i$ out of the neighborhood $\mathcal{A}_i = \{\textbf{x}| d_\textbf{M}(\textbf{x}_i,\textbf{x})< d_i^{max}\}$. This is consistent with the goal of the LMNN. Since the \emph{Lemma 3} proves the model described in Eq.(\ref{optim_continous2}) is convex when $\gamma_1 <0$, we can claim that the model in Eq.(\ref{optim_continous2}) is an improvement of LMNN.\\$\square$\\
\indent Although LANML ($\gamma_1 < 0$) serves for the same goal of LMNN, the searching space of LANML is much boarder than that of LMNN. In LMNN, all samples are considered equally. However, in the LANML, only the samples near the decision boundary are considered to form the constraints of the model. Therefore, the LANML ($\gamma_1\leq 0$) has much boarder searching space than that of LMNN.\\
\indent According to the \textbf{{Lemma 1}}, to keep the convexity of LANML, the number of separable similar samples should be larger than $\vert \mathcal{S}_i\vert/2$. Thus, half of the samples in $\mathcal{S}_i$ of LANML should be selected as the separable similar samples in advance with some given metrics. Thus, our method is superior to LMNN, because the ability of LANML to adaptively select the separable samples can fine-tune the selection of the similarity set $\mathcal{S}_i$. Therefore, LANML can avoid the situation that some useful samples would be ignored under the given metrics in LMNN. This is graphically shown in Fig. \ref{lmnn2_la}.
\begin{figure}[t]
 \centering
  \includegraphics[width=0.85\linewidth]{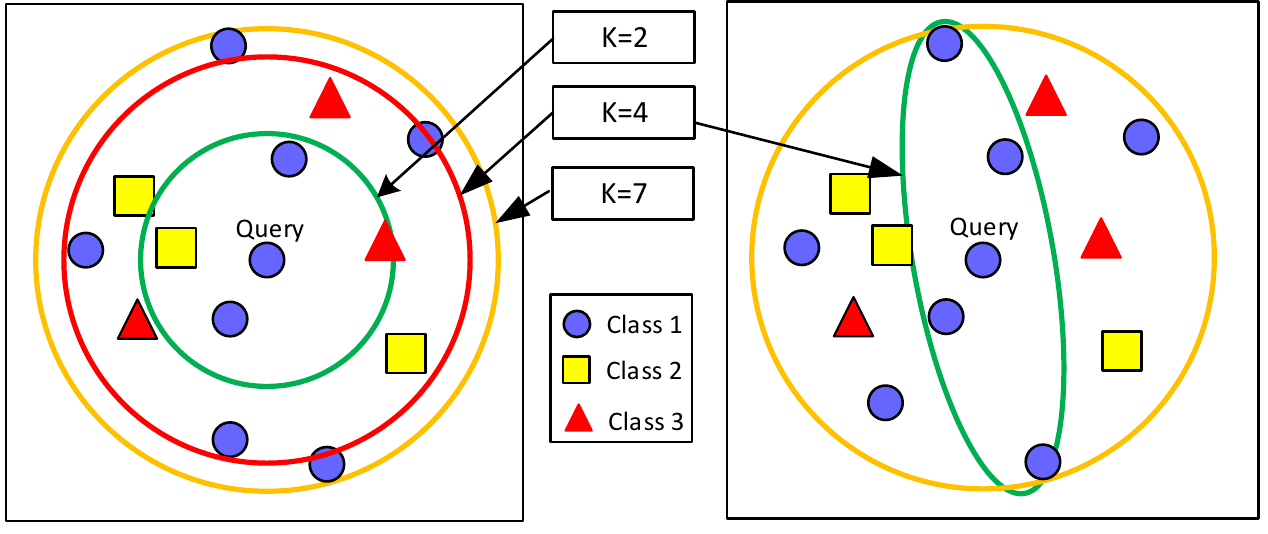}
  \vspace{-5pt}\caption{
  The left figure shows the cases of selecting target neighbors in LMNN under the Euclidean metric. The right figure shows the separable similar samples selected by LANML ($\gamma_1 <0$). Obviously, LANML ($\gamma_1 <0$) can select more separable samples than LMNN from the same similar sample set, and utilize more discriminant information to learn the metric.}
  \label{lmnn2_la}
\end{figure}
\begin{figure}[t]
 \centering
  \includegraphics[width=0.5\linewidth]{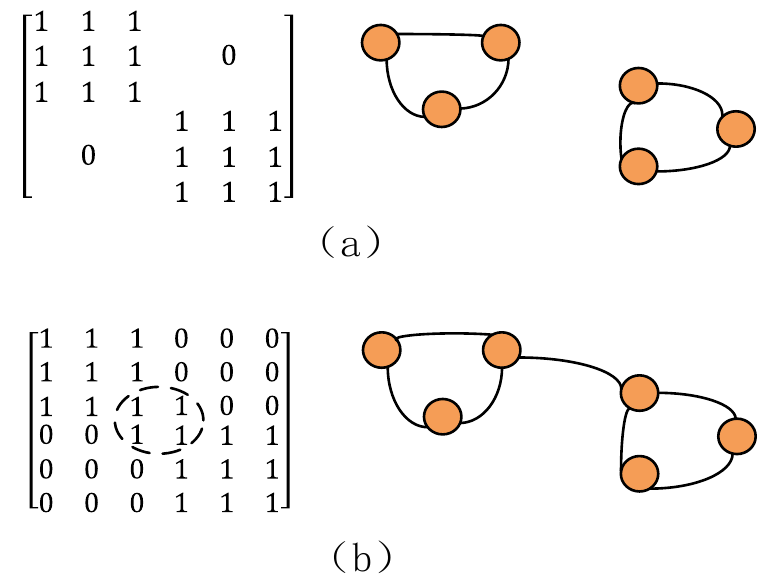}
  \vspace{-5pt}\caption{The illustration of how the adjacency matrix affects the cluster number. In (a), the adjacency matrix has two diagonal blocks, and the samples lie within two clusters. In (b), the elements $A_{34}$ and $A_{43}$ are not zero and the whole adjacency matrix becomes a diagonal block. Thus, the samples lie within one cluster. }
  \label{fig_neb2}
\end{figure}
\subsubsection{The Connection between LANML and NCA}
\label{pnca}
\textbf{Proposition 4:} \emph{When we set parameters $\gamma_1 =\gamma_2 = 1$ and the loss function $\ell(x)=x$, the LANML described in Eq.(\ref{optim_continous2}) is equivalent to the neighbourhood components analysis (NCA).}\\\vspace{-10pt}\\
\emph{Proof:} By setting the loss function as $\ell(x) = x$, the objective function of LANML for each query $\textbf{x}_i$ can be equivalently transformed as follows:
\begin{equation}\label{Eq_remark5}
\begin{split}
  &\min_{\textbf{M}\succeq 0} \ell(-\frac{1}{\gamma_1}\log[\frac{1}{\vert S_i\vert}\sum_{j \in \mathcal{S}_i} e^{-\gamma_1 d_\textbf{M}(\textbf{x}_i,\textbf{x}_j)}] +\\
  & \quad\quad\quad\quad\quad\quad\quad\frac{1}{\gamma_2}\log[\frac{1}{\vert D_i\vert}\sum_{j \in {\mathcal{D}}_{i}} e^{-\gamma_2d_\textbf{M}(\textbf{x}_i,\textbf{x}_j)}])\\
\end{split}
\end{equation}
\vspace{-6pt}
%Obviously, Eq.(\ref{Eq_remark5}) can be equivalently transformed in following formulation:
\begin{equation}\label{Eq_remark511}
\begin{split}
\Leftrightarrow& \max_{\gamma_2\textbf{M}\succeq 0}\frac{\sum_{j \in {\mathcal{D}}_{i}} e^{-d_{\gamma_2\textbf{M}}(\textbf{x}_i,\textbf{x}_j)}}{(\sum_{j \in \mathcal{S}_{i}} e^{-\alpha d_{\gamma_2\textbf{M}}(\textbf{x}_i,\textbf{x}_j)})^{1/\alpha}}, \alpha = \frac{\gamma_1}{\gamma_2}\\
\Leftrightarrow & \max_{\textbf{M}\succeq 0}\frac{(\sum_{j \in \mathcal{S}_{i}} e^{-\alpha d_\textbf{M}(\textbf{x}_i,\textbf{x}_j)})^{1/\alpha}}{{\sum_{j \in {\mathcal{D}}_{i}} e^{-d_\textbf{M}(\textbf{x}_i,\textbf{x}_j)}}+{(\sum_{j \in \mathcal{S}_{i}} e^{-\alpha d_\textbf{M}(\textbf{x}_i,\textbf{x}_j)})^{1/\alpha}}}
\end{split}
\end{equation}
where $\alpha$ in Eq.(\ref{Eq_remark511}) is introduced by the fact $\gamma_2d_\textbf{M}(\textbf{x}_i,\textbf{x}_j) = d_{(\gamma_2\textbf{M})}(\textbf{x}_i,\textbf{x}_j)$. Let $\mathcal{S}_i$ represent the set of all the samples in the class ${y_i}$ except $\textbf{x}_i$, and $\mathcal{D}_i$ represent the set of all the samples in the classes different from $y_i$. When we set $\alpha = 1$, the objective function in the last formulation presented in Eq.(\ref{Eq_remark511}) becomes the $p_i$ presented in the \emph{Eq.(2)} in \cite{Goldberger2004Neighbourhood}, which is the probability of $\textbf{x}_i$ being classified correctly. Therefore, the NCA is a special case of LANML. $\square$\\
 \indent Actually, the objective function in the last optimization problem in Eq.(\ref{Eq_remark511}) can be seen as a parameterized probability of $\textbf{x}_i$ being classified correctly, so we call it as the parameterized neighbourhood components analysis (PNCA). The parameter $\alpha$ plays the rule to balance the numbers of inseparable samples in the sets $\mathcal{S}_{i}$ and $\mathcal{D}_i$.
\vspace{-5pt}\section{Learning Deep Feature Embedding}
\subsection{Formulation}
\indent Let $f_{\theta}(\textbf{x}_i)$ be a deep neural network parameterized with $\theta$ and the distance function be $d_\theta(\textbf{x}_i,\textbf{x}_j)= \vert f_{\theta}(\textbf{x}_i)-f_{\theta}(\textbf{x}_j)\vert$, our model becomes the deep metric learning algorithm which is always solved by the statistic gradient descent (SGD) method needing to split the training data into several small parts called as the mini-batches. Commonly, in different mini-batches, the samples from the same class may have different distributions due to the sampling deviation. This may reduce the stability of the radius of the determined neighborhood, thus resulting in a poor performance. That is because, the average value of $K$ smallest numbers (or the largest numbers) obtained by the \emph{log-exp mean function} $b(\gamma)$ is sensitive to the distribution of the number series. For example, a fixed $\gamma$ may correspond to different values of $K$ for different mini-batches sampled from the same class. \\
\indent To alleviate this problem, we modify the constraint stated in Eq.(\ref{constaint2}) by introducing two constants $\lambda_1$ and $\lambda_2$ to form the radiuses of the neighborhoods $\mathcal{A}_i$ and $\mathcal{B}_i$, respectively. As a result, the optimization problem of the deep adaptive neighborhood metric learning (DANML) is obtained as follows:
  \begin{equation}\label{optim_continous_deep}
  \small
 \begin{split}
  \min_{\theta} \sum_{i=1}^N & \ell\left(-\frac{1}{\gamma_1} \log[\frac{1}{\vert\mathcal{S}_i\vert\!+\!\!1}(e^{-\gamma_1 \lambda_1}\!\!+\!\!\!\!\sum_{j\in\mathcal{S}_i}{e^{-\gamma_1 d_{{\theta}}(\textbf{x}_i,\textbf{x}_j)}})]\right.\\
  & \left. +\frac{1}{\gamma_2} \log[\frac{1}{\vert\mathcal{D}_i\vert\!\!+\!\!1}(e^{-\gamma_2\lambda_2}+\!\!\!\!\sum_{l\in\mathcal{D}_i}{e^{-\gamma_2d_{{\theta}}(\textbf{x}_i,\textbf{x}_l)}})]\right)
 \end{split}
 \end{equation}
 where $\gamma_1 < 0$ and $\gamma_2 > 0$. The reason why we set the parameter $\gamma_1 < 0$ is presented in the following proposition.\\\vspace{-5pt}\\
 \indent \textbf{ Proposition 5}: \emph{When $\gamma_1 < 0$, the learned features of each class only lie within one cluster.}\\\vspace{-5pt}\\
 \indent \emph{Proof:} We adopt the technique of graph partition to prove this proposition. Suppose $\textbf{x}_i$ and $\textbf{x}_j$ are two samples in the $c$-th class in which the adjacency matrix of samples is denoted as $\textbf{A}^c$. Considering two similar samples $\textbf{x}_i$ and $\textbf{x}_j$, the triplet constraint $d_\theta(\textbf{x}_i,\textbf{x}_j) <d_\theta(\textbf{x}_i,\textbf{x}_l)$ means there is a must-link between samples $\textbf{x}_i$ and $\textbf{x}_j$. As a result, the $ij$-th element of $\textbf{A}^c$ is set as $\textbf{A}^c_{ij} =1$ otherwise $0$. According to the \textbf{{Theorem 1}} in \cite{nie2016constrained}, when the samples from the same class are in one cluster, the rank of Laplace matrix of $\textbf{A}^c$ is $n_c-1$, where $n_c$ is the number of the samples in the $c$-th class. In this case, there is only one diagonal block in $\textbf{A}^c$ by permutating $\textbf{A}^c$'s rows and columns to transform it into a diagonal block matrix. To fulfill this goal, the number of non-zero elements of $\textbf{A}^c$ in each column (or row) should be larger than $n_c/2$. The illustration is shown in Fig. \ref{fig_neb2}. Thus, when $\gamma_1<0$, the number of separable samples in $\mathcal{A}_i$ is larger than $\vert \mathcal{S}_i\vert/2$.\\
 $\Box$\\
 \indent Then, we explain the meanings of $\lambda_1$ and $\lambda_2$. Let us denote $r_i=-\frac{1}{\gamma_1} \log[\frac{1}{\vert\mathcal{S}_i\vert+1}(e^{-\gamma_1 \lambda_1}+\sum_{j\in\mathcal{S}_i}{e^{-\gamma_1 d_{{\theta}}(\textbf{x}_i,\textbf{x}_j)}})]$ and $\mathcal{Q}_i^s = \{d_\theta(\textbf{x}_i,\textbf{x}_j)|{j\in\mathcal{S}_i} \}$. So $r_i$ is the average value of the $K_1$ largest values in $\mathcal{Q}_i^s\cup \{\lambda_1\}$. The value of $\lambda_1$ should be larger than the $K_1$-th largest value in $\mathcal{Q}^s_i$, or its impaction would be ignored. Thus, $r_i = \frac{\lambda_1 + s^r}{K_1}$, where $s^r$ is the sum of the $(K_1-1)$ largest values in $\mathcal{Q}^s_i$. Since different mini-batches share the same $\lambda_1$, it makes the estimated radius of the neighborhood more stable. Therefore, the values of $\lambda_1$ and  $\lambda_2$ are the average value of the radius of the corresponding neighborhoods of all mini-batches.
\begin{figure}[t]
 \centering
\includegraphics[width=0.55\linewidth]{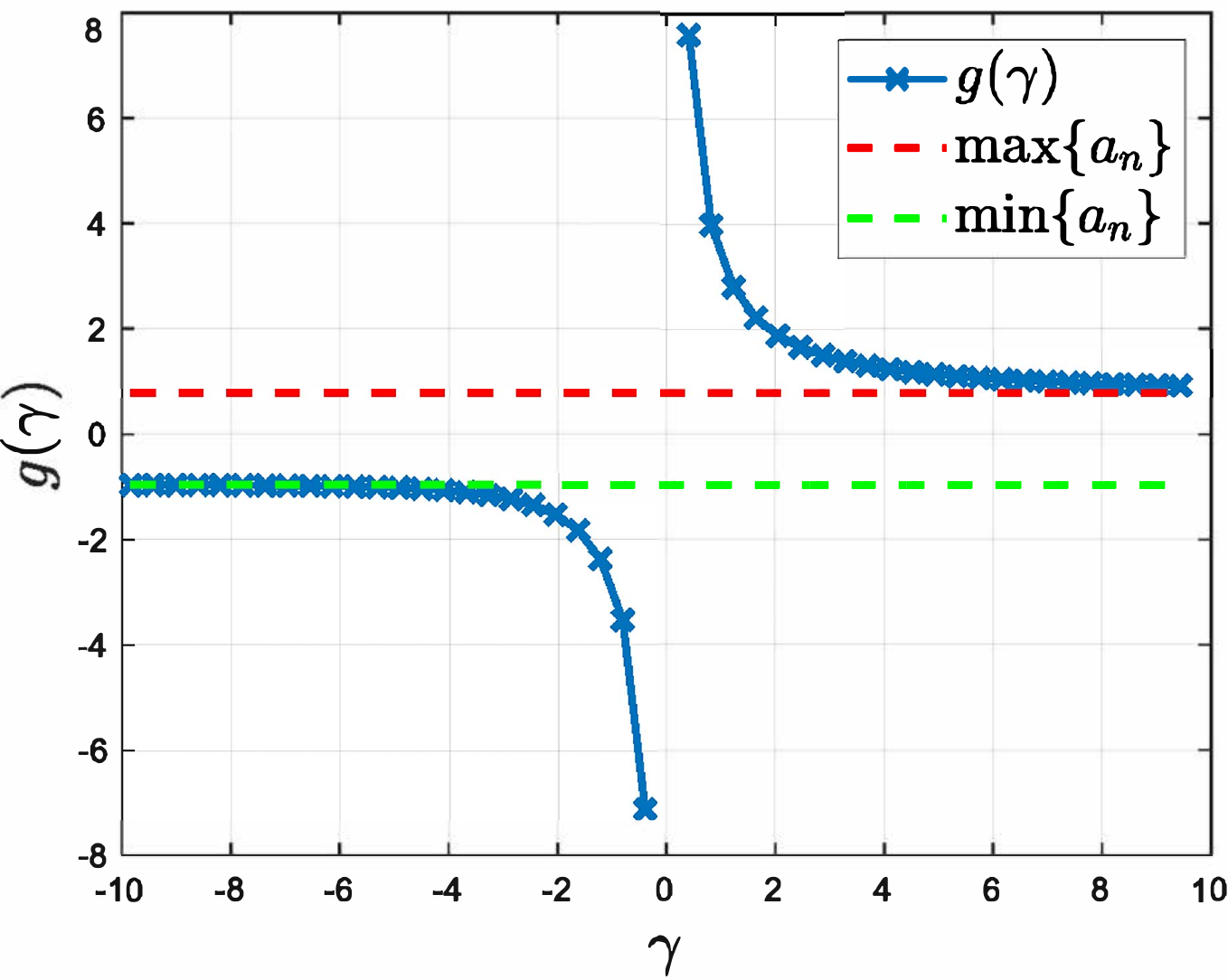}
\vspace{-5pt}\caption{\emph{The curve line of $g(\gamma) = \frac{1}{\gamma}log(\sum_{i=1}^ne^{\gamma a_i})$. When $\gamma>0$ ($\gamma<0$), $g(\gamma)$ is larger (smaller) than the maximal (minimal) value in the number series $\{a_1,a_2,\cdots,a_n\}$.}}
\label{fig5_runing1}
\end{figure}
\vspace{-5pt}\subsection{Review Existing Deep Metric Learning Methods from the Perspective of Neighborhood}
\subsubsection{Improved Lifted Structure Embedding}
The objective of the improved Lifted structure loss \cite{oh2016deep,dfense_triplet} is presented as follows:
\begin{equation}\label{imp_lift}
\small
\begin{split}
L(\textbf{X};f) =& \sum_{i=1}^N[\log(\sum_{y_k=y_i}exp(-S_{ik})) \\
&\quad\quad\quad\quad+ \log(\sum_{y_k\neq y_i}exp(S_{ik}))+ m]_+
\end{split}
\end{equation}
where $[-x+m]_+$ represents the hinge loss function with margin $m$. $S_{ik}$ is the similarity function between $\textbf{x}_i$ and $\textbf{x}_k$, and $y_i$ is the label of the $i$-th sample. \\
\indent The loss function in Eq.(\ref{imp_lift}) is constructed by utilizing the function $g(\gamma) = \frac{1}{\gamma}log(\sum_{i=1}^ne^{\gamma a_i})$ with $\gamma = 1$. The curve of $g(\gamma)$ is shown in Fig. \ref{fig5_runing1}. Suppose $S_i^{min} = \min_k\{S_{ik}|y_k=y_i\}$ and $D_i^{max} = \max_k\{ S_{ik}|y_k \neq y_i\}$. There is $-log(\sum_{y_k=y_i}exp(-S_{ik}))<S_i^{min}$ and $ log(\sum_{y_k\neq y_i}exp(S_{ik})) > S_i^{max}$. \\
\indent Therefore, the loss function in Eq.(\ref{imp_lift}) is a continuous proxy of the Triplet loss \cite{Schroff2015FaceNet} which punishes the samples violating the following equation:
\begin{equation}\label{lft_eq}
S_i^{min} > D_i^{max} -m
\end{equation}
\vspace{2pt}
\textbf{Proposition 6}: \emph{When $\gamma_1\rightarrow +\infty$ and $\gamma_2 \rightarrow -\infty$, the improved Lifted structure loss and Triplet loss are the special cases of our DANML.}\\\vspace{-5pt}\\
\indent This is easy to prove because the proposed DANML also punishes the constraint in Eq.(\ref{lft_eq}) with $\gamma_1\rightarrow +\infty$ and $\gamma_2 \rightarrow -\infty$. \\
\indent  Similar to Triplet loss, the model of improved Lifted structure embedding easily encounters inseparable problem. Besides, the performance of the improved Lifted structure loss is also reduced due to the random selection of min-batches. Therefore, we make a further improvement on it as follows:
\begin{equation}\label{imp_lift1}
\small
\begin{split}
L(\textbf{X};f) =& \sum_{i=1}^m[\frac{1}{\gamma_1}\log(exp(-\gamma_1\lambda_1)\!\!+\!\!\!\!\!\sum_{y_k=y_i}exp(-\gamma_1S_{ik}))\\
+ \frac{1}{\gamma_2}&\log(exp(\gamma_2\lambda_2)+\sum_{y_k\neq y_i}exp(\gamma_2S_{ik})) + m]_+
\end{split}
\end{equation}
where $\gamma_1$ and $\gamma_2$ are parameters to control the radiuses of neighborhoods $\mathcal{A}_i$ and $\mathcal{B}_i$, $[m-x]_+$ is the hinge loss with margin $m$, and $\lambda_1$ and $\lambda_2$ are used to reduce the turbulence of radiuses caused by the mini-batches constructed randomly.
\vspace{-10pt}\subsubsection{Multi-Similarity Loss}
\indent The objective function of multi-similarity loss \cite{wang2019multi} is calculated as:
\begin{equation}\label{ms_loss}
\small
\begin{split}
 L({D_{ij}}) = &\sum_{i=1}^N(\frac{1}{\alpha}log(1+\sum_{j \in \mathcal{P}_i}exp(\alpha(D_{ij}-m)))\\
 &+ \frac{1}{\beta}log(1+\sum_{k \in \mathcal{N}_i}exp(\beta(m-D_{ik}))))
\end{split}
\end{equation}
where $D_{ij}$ is the cosine of the angle between the $i$-th and the $j$-th embeddings. $\alpha$, $\beta$ and $m$ are predefined hyper-parameters to well control the weights for different pairs.\\\vspace{-5pt}\\
\indent\textbf{Proposition 7}: \emph{The multi-similarity loss is a special case of DANML when the loss function in Eq.(\ref{optim_continous_deep}) is set as $loss(x) = x$, and the parameters $\lambda_1$ and $\lambda_2$ are set as $\lambda_1 = \lambda_2= m$.}\\\vspace{-5pt}\\
\indent\emph{Proof:} To prove the \emph{Proposition 7}, we equivalently transform the multi-similarity loss as follows:
\begin{equation}\label{ms_loss_tr}
\small
\begin{split}
& L({D_{ij}}) =\sum_{i=1}^N\frac{1}{\alpha}log[\frac{exp(\alpha m)+\sum_{j \in \mathcal{P}_i}exp(\alpha(D_{ij}))}{\vert \mathcal{P}_i\vert+1}] \\
&\quad\quad+\frac{1}{\beta}log[\frac{exp(\beta(-m))+\sum_{k \in \mathcal{N}_i}exp(\beta(-D_{ik}))}{\vert \mathcal{N}_i\vert+1}]\\
\end{split}
\end{equation}
\indent Let us set $loss(x) = x$, $\lambda_1 = \lambda_2= m$ in Eq.(\ref{optim_continous_deep}), and the model of DANML becomes the multi-similarity loss. $\Box$\\
 \indent In this way, we can interpret the meaning of the parameter $m$ used in the multi-similarity loss as the quantity to reduce the turbulence of each similarity set $\mathcal{S}_i$ and dissimilarity set $\mathcal{D}_i$ caused by the construction of mini-batches. However, our method gives the distinct meaning of parameters $\gamma_1$ and $\gamma_2$, while the multi-similarity loss does not.\\
\vspace{-15pt}\subsubsection{N-pair Loss}
\vspace{-0pt}N-pair Loss \cite{sohn2016improved} is designed to overcome the shortcoming of Triplet loss which only pulls one positive point while pushing a negative one simultaneously. The formulation of N-pair loss is:
\begin{equation}\label{n_pair_loss}
\small
L(\{(\textbf{x}_i,\textbf{x}_i^+)\}_{i=1}^N;{f}) \!=\! \frac{1}{N}\sum_{i=1}^N log\{1\!+\!\sum_{j\neq i} e^{(\textbf{f}_i^T\textbf{f}_j^+ \!\!-\!\!\textbf{f}^T_i\textbf{f}^+_i)}\}
\end{equation}
where $\textbf{f}_i = f(\textbf{x}_i)$. $\{(\textbf{x}_i,\textbf{x}_i^+)\}_{i=1}^N$ are $N$ pairs of examples from $N$ different classes, i.e., $y_i\neq y_j$, $\forall i \neq j$. Here, $\textbf{x}_i$ and $\textbf{x}_i^+$ are the query and the positive example, respectively..\\
\indent We explain the geometry meaning of N-pair loss by utilizing the conclusion of \textbf{{Theorem 3}}. The objective function in Eq. (\ref{n_pair_loss}) can be equivalently transformed as follows:
\begin{equation}\label{n_pair_loss1}
\small
\begin{split}
L(\{(\textbf{x}_i,\textbf{x}_i^+)\}_{i=1}^N;f) \!=\!&\frac{1}{N}(\sum_{i=1}^N\!\!log(\frac{e^{\textbf{f}^T_i\textbf{f}^+_i}
\!+\!\sum_{j\neq i} e^{\textbf{f}_i^T\textbf{f}_j^+}}{N+1})\\
&-\textbf{f}^T_i\textbf{f}^+_i)
\end{split}
\end{equation}
Suppose $s^{[k]}$ is the $k$-th smallest element in $\{\textbf{f}^T_i\textbf{f}^+_i,\textbf{f}^T_i\textbf{f}^+_1,\cdots,\textbf{f}^T\textbf{f}^+_{N-1}\}$, according to the \textbf{Theorem 3}, there is an integer $K$ to hold $\frac{1}{K}\sum_{k=1}^K s^{[N-k]}= log(\frac{e^{\textbf{f}^T_i\textbf{f}^+_i}+\sum_{j\neq i} e^{\textbf{f}_i^T\textbf{f}_j^+}}{N+1})$. Thus, the essence of the N-pair loss is to punish the samples violating the following constraint:
\begin{equation}\label{N_pair_cons}
\small
\frac{1}{K}\sum_{k=1}^K s^{[N-k]}<\textbf{f}^T_i\textbf{f}^+_i
\end{equation}
Eq. (\ref{N_pair_cons}) means to push all the similar samples in the neighborhood $\mathcal{B}_i$ defined in Eq.(\ref{neighbor_di}) with the radius $\frac{1}{K}\sum_{k=1}^K s^{[N-k]}$. Thus, N-pairs loss discards the inseparable negative samples with the similarity value larger than $\frac{1}{K}\sum_{k=1}^K s^{[N-k]}$.\\
\indent According to the geometrical meaning of neighborhood, we make an improvement on the N-pairs loss presented as follows:
\begin{equation}\label{n_pair_loss}
\small
L(\{(\textbf{x}_i,\textbf{x}_i^+)\}_{i=1}^N;{f}) \!\!=\!\! \frac{1}{N\gamma}\sum_{i=1}^N log ( 1\!\!+\!\!\sum_{j\neq i}e^{\gamma(-\lambda+\textbf{f}_i^T\textbf{f}_j^+ \!\!-\!\!\textbf{f}^T_i\textbf{f}^+_i)})
\end{equation}
where $\gamma$ is the parameter to control the radius of neighborhood $\mathcal{B}_i$, and $\lambda$ is used to reduce the turbulence of radius of $\mathcal{B}_i$ caused by the random construction of mini-batches.
%\vspace{-14pt}\subsubsection{Circle Loss}
%\indent The Circle loss is defined as $\mathcal{L}_{circle} = log[1+\sum_{i=1}^K\sum_{j=1}^L exp(\gamma(\alpha_n^j(s_n^j-\Delta_n)-\alpha_p^i(s_p^i-\Delta_p))]$, where $s_p^i$ and $s_n^j$ are distances of the $i$-th positive and $j$-th negative samples to the query sample $\textbf{x}$, and $\alpha_n^j = [s_n^j - O_n]_+$, $\alpha_p^j = [O_p -s_p^i]_+$. The $O_n$, $O_p$, $\gamma$, $\Delta_n$, $\Delta_p$ are five parameters needed to tune. The Circle loss is obtained by generalizing the following loss function:
%\begin{equation}\label{general_loss}
%\mathcal{L}_{circle} = log[1+\sum_{i=1}^K\sum_{j=1}^L exp(\gamma((s_n^j-s_p^i+m)]
%\end{equation}
%where $m$ is the margin. Let the symbol $d^{[-k]}$ represent the $k$-th largest value in set $\{s_n^j-s_p^i+m|i=1\cdots K,j=1\cdots L\}$. According to the \textbf{Theorem 2}, there is a positive integral $P$ to let $\frac{1}{\gamma}log[1+\sum_{i=1}^K\sum_{j=1}^L exp(\gamma((s_n^j-s_p^i+m)] = log(LK+1)+ \frac{1}{P}\sum_{k=1}^P d^{[-k]}$. Thus, the goal of Eq.(\ref{general_loss}) is to minimize $\frac{1}{P}\sum_{k=1}^P d^{[-k]}$. Since $s_n^j-s_p^i+m$ corresponds to the triplet constraint $s_n^j>s_p^i+m$, when the value of $s_n^j-s_p^i+m$ is larger, the more likely the triplet is inseparable. Thus, by setting an appropriate $\gamma$, minimizing $\frac{1}{P}\sum_{k=1}^P d^{[-k]}$ can ignore those inseparable triplets. This is why the Circle Loss works.
%
\begin{table}[!htbp]
\vspace{-7pt}\renewcommand{\arraystretch}{0.7} % 行间距
\setlength{\tabcolsep}{12pt}  % 列间距
\centering
\caption{The Details of Several Datasets.}
\label{data_set}
\footnotesize
\vspace{-6pt}\begin{tabular}{|c||c|c|c|}
\toprule
Data set & \# Classes& \#Examples& \#Features\\
\midrule
Australian&2&690&14\\
Cars&2&392&8\\
Ecoli&8&336&343\\
German &2&1,000&20\\
Glass&6&214&9\\
Iris&3&150&4\\
Isolet&2&1,560&617\\
Monk1&2&432&6\\
Solar&6&323&12\\
Vehicle&4&846&18\\
Wine&3&178&13\\
Pendigits&10&10,992&16\\
Coil20&20&1,440&1024\\
Letter&26&20,000&16\\
Usps&10&9,298&256\\
\bottomrule
\end{tabular}
\end{table}
%
%
%
%%%%%%%%%%%%%%%%%%%%%%%%%%%%%%%%%%%%%%%%%%%%%%%%%%%%%%
%
%%%%%%%%%%%%%%%%%%%%%%%%%%%%%%%%%%%%%%%%%%%%%%%%%%%%%%%%%
\vspace{-3pt}\section{Numerical Experiments}
\subsection{Evaluation of LANML for Mahanoibis Distance Metric Learning}
\subsubsection{Data Set Description and Experimental Settings}
We evaluate the proposed LANML on 15 data sets which are widely adopted to evaluate the performance of machine learning algorithms. All of those data sets come from the UCI Machine learning Repository\footnote{Available at http://archive.ics.uci.edu/ml/datasets.html} and LibSVM\footnote{https://www.csie.ntu.edu.tw/$\sim$cjlin/libsvm/}. Since the feature values in some data sets are very large, we normalize them by subtracting the mean and dividing the standard deviation for each feature. The scales of those data sets range from 178 to 20000. Their dimensions vary from 4 to 1024, and the number of classes changes from $2$ to $26$. The details of the data sets are presented in Table {\ref{data_set}}. For the data sets whose feature numbers are larger than 150, we utilize principal components analysis (PCA) to reduce the number of their dimensions to $150$.\\
\indent As discussed in section {\ref{relationship_ann_other}}, by setting $\gamma_1>0$ and $\gamma_1<0$, the proposed LANML would become different methods, respectively. We use the symbols LANML$^{+}$ and LANML$^{-}$ to represent the cases of LANML with $\gamma_1 >0$ and $\gamma_1<0$, respectively. We set the regularization term as $\Omega(\textbf{M}) = \frac{1}{N\vert\mathcal{S}_i\vert} \sum_{i=1}^N\sum_{j\in \mathcal{S}_i} d_{\textbf{M}}(\textbf{x}_i,\textbf{x}_j)$, and loss function as hinge loss. Since the PNCA presented in Eq.(\ref{Eq_remark511}) is also derived from our proposed method, we also evaluate its performance in this section.
\subsubsection{Comparison of Classification Accuracy}
\indent In this section, we evaluate the proposed methods on 15 data sets. For those data sets, each of them are split into $70\%/30\%$ partition for training and testing for 30 times, and the average classification results are reported.\\
\indent We adopt $8$ state-of-the-art methods as comparison. They are large margin nearest neighbor (LMNN) \cite{weinberger2005distance}, information theoretic metric learning (ITML) \cite{davis2007information}, local distance metric learning (LDML) \cite{guillaumin2009you}, sparse component metric learning (SCML) \cite{scml14}, BoostMetric \cite{Shen2009Positive}, neighbourhood components analysis (NCA) \cite{Goldberger2004Neighbourhood}, geometric mean metric learning (GMM) \cite{zadeh2016geometric} and regressive virtual metric learning (RVML) \cite{perrot2015regressive}, etc. Our methods include LANML$^{+}$, LANML$^{-}$ and PNCA.\\
 \indent In LANML$^{-}$, the similarity set $\mathcal{S}_i$ is constructed by selecting $10$ nearest neighbors of $\textbf{x}_i$ from the class $y_i$ under the Euclidean metric, and the dissimilarity set $\mathcal{D}_i$ is constructed by all of the samples with different labels from $y_i$. In LANML$^{+}$ and PNCA, the similarity set of $\textbf{x}_i$ is constructed by all of the samples in the class $y_i$ except for $\textbf{x}_i$. The dissimilarity set is constructed by all of the samples in the classes different from $y_i$. The parameter $\lambda$ in LANML$^{+}$ and LANML$^{-}$ is tuned in the grid of $\{0.1,0.3,\cdots,1.5\}$. The $\gamma_1$ in LANML$^{+}$ is tuned in $\{2^{-5},2^{-4.5},\cdots,2^{5}\}$. The $\alpha$ in PNCA is tuned in $\{2^{-8},2^{-7},\cdots,2^{10}\}$, and the $\gamma_1$ in LANML$^{-}$ is tuned at the searching grid of $\{-2^{-5},-2^{-4.5},\cdots,-2^{5}\}$. The $\gamma_2$ in LANML$^{+}$ and LANML$^{-}$ is tuned in $\{2^{-5},2^{-4.5},\cdots,2^{5}\}$. Since LANML$^{+}$ and PNCA are non-convex optimization problems, we set the initial searching point as $\textbf{M}_0 = \frac{\textbf{I}}{\sqrt{N}}$, where $\textbf{I}$ is the identity matrix. \\
\indent In LMNN, $\lambda$ is tuned at the searching grid of $\{0.1,0.2,\cdots,0.9\}$, the target neighbors' number is searched in the grid of $\{4,\cdots,10\}$. For GMML, the parameter $t$ is tuned in the grid of $\{0.1,0.2,\cdots,0.9\}$. For ITML, the parameter $\gamma$ is tuned in the grid of $\{0.25,0.5,0.7,0.9\}$. All of other parameters are set as default. Those tuned parameters are determined with 5-fold cross validation. After the metric learning step, we report the best results output by $K$-NN with $K\in \{1,2,\cdots,40\}$. The result are shown in Table \ref{ta1_result_cur}. As seen in Table \ref{ta1_result_cur}, the following conclusions are made:
\begin{itemize}
  \item The proposed methods, LANML$^{+}$, LANML$^{-}$ and PNCA, have achieved better results compared with most of the comparison methods. This demonstrates the superiority of the proposed methods.
  %\item On some data sets, LANML$^{+}$ has obtained best results, and on other data sets, LANML$^{-}$ has obtained the best results. This is very similar to the comparison between LMNN and NCA. Considering that the LANML$^{+}$ is a non-convex model whose results rely on the initial searching points. With an appropriate initial searching point, the performance of LANML$^{+}$ may be further improved. However, the appropriate initial point is hard to find. Similarly, the performance of LANML$^{-}$ depends on the number of the samples of $\mathcal{S}_i$, but the appropriate number $\vert \mathcal{S}_i\vert$ is much easier to find compared with the initial point in LANML$^{+}$.
  \item LANML$^{-}$ has achieved better performance than LMNN. That is because LANML$^{-}$ has the ability to fine-tune the selection of target neighbors. As a result, LANML$^{-}$ has a larger searching space than that of LMNN.
  \item PNCA has achieved better results than NCA on most of the data sets. That may be because the parameter $\alpha$ could adjust the model to suit the data set better.
\end{itemize}
\begin{table*}[!htbp]
\renewcommand{\arraystretch}{1} % 行间距
\setlength{\tabcolsep}{2pt}  % 列间距
\centering
\caption{Comparison of Different Methods on 15 Data Sets.}
\label{ta1_result_cur}
\footnotesize
\vspace{-6pt}\begin{tabular}{ccccccccc|ccc}
\toprule
&\multicolumn{8}{c|}{Baselines}&\multicolumn{3}{c}{Our Methods}\\
\midrule
Dataset &NCA&LMNN&ITML&LDML&SCML&RVML&GMML&BoostMetric&PNCA&LANML$^{+}$&LANML$^{-}$\\
\midrule
Australian&71.12$\pm$2.14&71.16$\pm$2.69&67.39$\pm$2.42&70.32$\pm$2.22&70.26$\pm$1.98&73.12$\pm$2.11&\textbf{85.94$\pm$2.32}&72.42$\pm$2.13&{79.15$\pm$2.42}&{83.68$\pm$2.35}&{80.84$\pm$2.27}\\
cars &80.96$\pm$2.21&83.33$\pm$1.94&81.38$\pm$2.12&80.16$\pm$2.47&83.11$\pm$2.28&82.68$\pm$2.51&84.91$\pm$2.26&84.16$\pm$2.17&{82.76$\pm$2.13}&\textbf{85.97$\pm$2.13}&{83.49$\pm$2.09}\\
Ecoli&76.32$\pm$1.59&79.15$\pm$1.43&81.52$\pm$1.27&81.22$\pm$1.79&80.17$\pm$1.93&81.32$\pm$1.45&76.22$\pm$1.24&77.47$\pm$1.63&81.54$\pm$1.72&{83.58$\pm$1.92}&\textbf{84.36$\pm$1.42}\\
German&67.23$\pm$2.41&78.51$\pm$2.21&74.51$\pm$2.65&77.31$\pm$2.49&75.22$\pm$2.37&74.14$\pm$2.55&71.62$\pm$2.04&77.11$\pm$2.29&{75.89$\pm$2.07}&\textbf{79.91$\pm$2.33}&{79.71$\pm$2.21}\\
Glass&70.09$\pm$1.34&71.43$\pm$1.56&65.88$\pm$1.37&71.11$\pm$1.29&66.32$\pm$1.54&71.21$\pm$1.57&62.61$\pm$1.61&70.22$\pm$1.76&{73.42$\pm$1.62}&{75.56$\pm$1.61}&\textbf{76.77$\pm$1.53}\\
Iris&95.76$\pm$1.89&96.11$\pm$1.96&96.67$\pm$2.01&96.21$\pm$2.16&95.22$\pm$2.32&96.11$\pm$2.22&97.47$\pm$2.18&96.23$\pm$2.04&{98.87$\pm$2.09}&\textbf{99.89$\pm$2.09}&{99.79$\pm$2.09}\\
Isolet&83.9$\pm$2.03&87.57$\pm$2.12&84.05$\pm$2.10& 85.17$\pm$2.17&86.28$\pm$1.98&88.06$\pm$2.32&82.62$\pm$2.19&86.38$\pm$1.82&    90.31$\pm$2.24&\textbf{93.85$\pm$1.95}&{88.38$\pm$2.03}\\
Monk1&83.42$\pm$1.72&86.27$\pm$1.88&86.84$\pm$1.64&86.11$\pm$1.82&86.64$\pm$1.88&84.34$\pm$1.76&89.16$\pm$1.73&85.43$\pm$1.62&85.62$\pm$1.59&\textbf{91.74$\pm$1.74}&{87.43$\pm$1.74}\\
Solar&68.21$\pm$2.38&70.11$\pm$2.34&62.12$\pm$2.41&65.12$\pm$2.25&65.22$\pm$2.18&66.33$\pm$2.49&64.05$\pm$2.53&63.61$\pm$2.41&71.52$\pm$2.21&{72.57$\pm$2.25}&\textbf{73.99$\pm$2.50}\\
Vehicle&71.22$\pm$2.22&73.96$\pm$2.18&68.79$\pm$2.19&72.21$\pm$2.26&72.91$\pm$2.52&70.12$\pm$2.24&78.15$\pm$2.17&72.18$\pm$2.19&72.49$\pm$2.15&{76.78$\pm$2.31}&\textbf{78.79$\pm$2.24}\\
Wine&87.63$\pm$2.24&89.14$\pm$2.12&89.32$\pm$2.13&89.41$\pm$2.32&88.34$\pm$2.39&90.12$\pm$2.11&86.32$\pm$2.22&90.13$\pm$2.32&92.27 $\pm$2.12&{97.15$\pm$2.13}&\textbf{98.15$\pm$2.19}\\
Pendigits&94.12$\pm$1.36&97.72$\pm$1.46&94.24$\pm$1.46&95.32$\pm$1.45&96.43$\pm$1.52&98.22$\pm$1.62&94.21$\pm$1.05&96.53$\pm$2.11&95.46$\pm$2.21&98.43$\pm$1.62&\textbf{98.54$\pm$1.29}\\
Coil20&94.32$\pm$2.01&95.21$\pm$2.12&94.17$\pm$2.16&93.25$\pm$2.21&{96.81$\pm$2.09}&96.52$\pm$1.96&93.42$\pm$2.21&93.26$\pm$1.95&94.72$\pm$2.15&97.47$\pm$2.18&\textbf{97.81$\pm$2.03}\\
Letter&93.74$\pm$2.41&95.63$\pm$2.47&93.83$\pm$2.44&94.42$\pm$2.21&93.31$\pm$2.31&\textbf{95.72$\pm$3.61}&94.12$\pm$2.68&95.11$\pm$2.45&94.26$\pm$2.31&{95.15$\pm$2.27}&95.49$\pm$2.27\\
USPS&92.45$\pm$2.21&94.52$\pm$2.31&91.07$\pm$2.12&92.14$\pm$2.05&93.51$\pm$2.26&92.72$\pm$2.62&94.32$\pm$2.31&95.24$\pm$2.11&93.57$\pm$2.15&95.21$\pm$2.06&\textbf{95.54$\pm$2.02}\\
\bottomrule
\end{tabular}
\end{table*}
\vspace{-0pt}\subsubsection{Running Time of the Proposed Method}
\vspace{-0pt}In this section, we compare the running times of the proposed methods with LMNN. Since the training procedures of LANML$^{+}$ and LANML$^{-}$ may have different iterations, we report the running times of them separately. For LANML$^{+}$ and LANML$^{-}$, both sets $\mathcal{S}_i$ and $\mathcal{D}_i$ of each inquiry sample $\textbf{x}_i$ are constructed to the largest volume. For LMNN, the number of target neighbors $K_t$ is an essential factor for affecting the running time, we report the running time of LMNN with $K_t = \{4,5,6,7\}$. The parameters of regularization terms of LANML$^+$, LANML$^-$ and LMNN are set as $1$, $1$ and $0.5$, respectively. The $\gamma_1$ in LANML$^{-}$ and LANML$^{+}$ are set as $-1$ and $1$, respectively. For the three methods, each algorithm runs $30$ times without any accelerating strategies, and the average running time is recorded. All methods are implemented by MATLAB2017 on Intel(R)Xeon(R)CPU X5650 @2.80GHz, memory 80GB.  The results are shown in Fig.\ref{fig5_runing}.\\
 \indent As seen from Fig. \ref{fig5_runing}, LANML$^{-}$ and LANML$^{+}$ run much faster than LMNN. Theoretically, the running time of LANML$^{-}$ (or LANML$^{+}$) may be the $\frac{1}{K_t}$ of that of LMNN, however, in practice the proposed method is faster than LMNN by an order of magnitude. That is because the hinge loss function penalizes the constraint involved in $N$ distance computations. When the hinge loss is not triggered, $N$ distance computations are removed. However, in LMNN one hinge loss only triggers two distance computations. Another possible reason is that the objective function of LANML is more smooth than that of LMNN, so the convergence speed of the proposed methods is faster than LMNN.\\
%\indent Besides, we can observe that LANML$^{-}$ runs faster than LANML$^{+}$ on most of the data sets, but the difference is not very significantly. That may be because LANML$^{-}$ is a convex problem which is easier to find the convergence point than LANML$^{+}$ which is a non-convex one.
\begin{figure}[t]
 \centering
\includegraphics[width=0.65\linewidth]{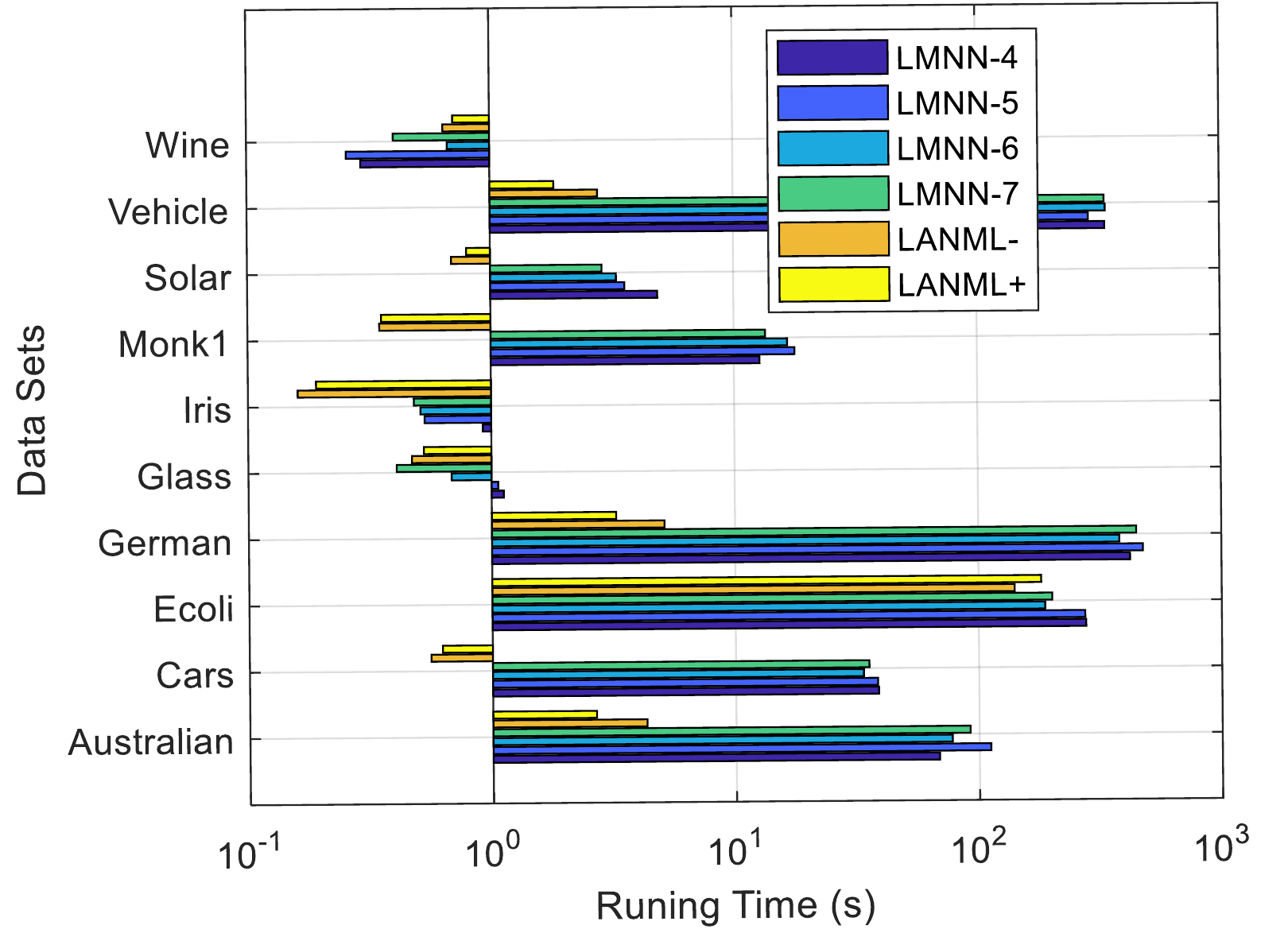}
\caption{The running times of different methods on different data sets.  The LMNN-4,  LMNN-5, LMNN-6, LMNN-7 represent LMNN with target neighbor number $K_t = \{4,5,6,7\}$, respectively. LANML$^{+}$ and LANML$^{-}$ represent LANML with parameter $\gamma_1 >0$ and $\gamma_2 <0$, respectively.}
\label{fig5_runing}
\end{figure}
\vspace{-10pt}
\vspace{-5pt}
\subsection{Evaluation of DNAML for Deep Metric Learning}
\indent In this section, we evaluate the DANML in two aspects. They are the ablation experiments and the comparison of different state-of-the-art algorithms, respectively.\\
%\begin{figure}[t]
% \centering
%\includegraphics[width=0.68\linewidth]{dimension.pdf}
%\caption{The effect of the embedding size on DANML (On the CUB-200-2011). We can find that the results at dimension $512$ and $1024$ are comparable.}
%\label{fig5_runing4}
%\end{figure}
\vspace{-17pt}\subsubsection{Experimental Settings}
\indent We implemented DANML by PyTorch on a singe Tesla V100 GPU with 32GB RAM. The loss function is selected as logistic loss. To fairly compare with previous works, we used the Inception network \cite{Ioffe2015Batch} with batch normalization pre-trained on ILSVRC2012-CLS \cite{ImageNet2015} to extract CNN features, and added a FC layer with the $L_2$ normalization as the feature embedding projector to keep all the features located on a sphere. Thus, the $D_{ij}$ was represented by negative cosine similarity function \cite{sohn2016improved}. Following \cite{wang2019multi}, we randomly crop all the images to $224\times 224$, and performed random horizontal mirroring for data augmentation. All experiments are trained by the Adm optimizer.\\
\indent Following previous experiments, CUB-200-2011 \cite{wah2011caltech}, Cars-196 \cite{krause20133d}, Stanford Online Products (SOP) \cite{oh2016deep} and In-Shop Clothes Retrieval (In-Shop) \cite{liu2016deepfashion} are adopted to evaluate the proposed method. The data split protocols are followed the one applied in  \cite{oh2016deep}.\\
%\indent We adopt the sample mining strategy used in \cite{wang2019multi} to select the similar and dissimilar samples having gradients with large amplitude for training. For every mini-batch, we randomly choose a certain number of classes, and then randomly sample $5$ instances for each class for all the data sets in all experiments. At last, all the methods are evaluated on image retrieval task by using the standard performance metric: Recall@K.\\
%\indent As CUB-200-2011 dataset consisting of 200 classes with 11,788 images, the first 100 classes with 5,864 images are used for training, and the 100 classes left with 5,924 images are for testing. For Cars-196 dataset having 16,185 images of cars with 196 categories, its first 98 categories are used for training, and the rest is used for testing. For the SOP dataset, we use 11,318 classes for training, and 11,316 classes for testing. For the In-shop dataset, we follow \cite{opitz2018deep} to use 3,997 classes with 25,882 images for training. The test set is partitioned into a query set with 14,218 images of 3,985 classes, and a gallery set having 3,985 classes with 12,612 images.\\
%\indent For the parameters in Eq.(\ref{optim_continous_deep}), we tune $\gamma_1 \in [1,3]$, $\gamma_2 = [25,35]$, $\lambda_1 \in [0.5,0.65]$, and $\lambda_2 = \lambda_1+ \sigma$ where $\sigma \in[0.01,0.05]$, respectively.
\indent For the parameters in Eq.(\ref{optim_continous_deep}), we tune $\gamma_1 \in \{1,2,3\}$, $\gamma_2 = \{25,30,35\}$, $\lambda_1 \in \{0.5,0.55,0.6,0.65\}$, and $\lambda_2 = \lambda_1+ \sigma$ where $\sigma \in\{ 0.01,0.02,0.03\}$, respectively. Besides, we adopt the sample mining strategy used in \cite{wang2019multi} to select the similar and dissimilar samples having gradients with large amplitude for training. For every mini-batch, we randomly choose a certain number of classes, and then randomly sample $5$ instances for each class for all the data sets in all experiments. At last, all the methods are evaluated on image retrieval task by using the standard performance metric: Recall@K.\\
\vspace{-13pt}\subsubsection{Ablation Study}
%\indent \textbf{Embedding Size:} By following \cite{sohn2016improved}, we explore how the embedding dimension affects the performance of DANML on the CUB-200-2011 dataset. We set the dimension range as $ \{64,128,256,512,1024,2048\}$ and results of Recall@1 are recorded. As shown in Fig. \ref{fig5_runing4}, with the embedding dimension growing, the performance is increased first and then reduced with a peak at $1024$. As the variance of the results at $1024$ and $512$ is not significant, we set the dimension as $512$ in the following experiments.\\
\indent \textbf{Improvement by adding parameters $(\lambda_1,\lambda_2)$ and $(\gamma_1,\gamma_2)$:} To validate the effectiveness of the improvement made from the perspective of neighborhood, we evaluate the performance of some state-of-the-art algorithms and their improved versions made in our paper on CUB-200-2011 dataset. The methods consist of Lifted structure loss, Prox-NCA, N-pairs loss, etc. The results are shown in Table \ref{ta_imporve}. We can find that, with the improvements, the performances of Lifted structure loss, Proxy-NCA, and N-pairs loss have increased with $1.7\%$, $3.5\%$, $3.2\%$ at $Recall@1$, respectively.
%Their relative improvement rates are about $3.76\%$, $7.11\%$, $7.33\%$, respectively.
 The improvement on Lifted structure loss is less than prox-NCA and N-pairs loss may be because the $(\gamma_1,\gamma_2)$ can not change the radius of neighborhoods too much like the other two methods. For DANML, we explore its performance influenced by parameters $(\lambda_1,\lambda_2)$. By adding $(\lambda_1,\lambda_2)$, its performance has increased $2.2\%$. Those outstanding results have validated the effectiveness of our neighborhood-based improvement.
\begin{table}[!htbp]
\vspace{0pt}\renewcommand{\arraystretch}{0.5} % 行间距
\setlength{\tabcolsep}{7pt}  % 列间距
\centering
\caption{Performance on the CUB-200-2011 of the three state-of-the-art methods and their improved versions with 512 dimension.}
\label{ta_imporve}
\footnotesize
%\small
\vspace{-6pt}\begin{tabular}{l|c|cccc}
\toprule
\multicolumn{2}{c|}{$Recall @ k$}&1&2&4&8\\
\midrule
\multirow{2}{2cm}{\tiny{Lifted structure loss$^{512}$}}&Original&45.4&58.4&69.5&79.5\\
&Improved&47.1&60.3&71.6&81.8\\
\hline
\multirow{2}{2cm}{Proxy-NCA$^{512}$}&Original&49.2&61.9&67.9&72.4\\
&Improved&52.7&65.4&68.3&75.7\\
\hline
\multirow{2}{2cm}{N-pairs loss$^{512}$}&Original&43.6&56.6&68.6&79.6\\
&Improved&46.8&60.7&72.6&83.8\\
\hline
\multirow{2}{2cm}{DANML$^{512}$}&Original&65.4&76.8&85.7&90.7\\
&Improved&67.6&79.1&88.2&93.4\\
\bottomrule
\end{tabular}
\end{table}
\vspace{-5pt}
\subsubsection{Comparison with State-of-the-Art Methods}
\indent In this section, we compare DANML with the state-of-the-arts on the CUB-200-2011 and Cars-196, In-shop and SOP datasets. Following the experimental protocol \cite{oh2016deep}, we report Recall@K with $K=\{1,2,4,\cdots,32\}$ for CUB-200-2011 and Cars-196; Following the work \cite{oh2016deep}, $K$ is set as $\{1,10,20,\cdots,50\}$ for In-shop dataset; For SOP, the $K$ is set as $\{1,10,100,1000\}$ \cite{opitz2018deep}. The results are shown in Tables \ref{ta1_result}-\ref{ta3_result}.\\
\indent As shown in Table \ref{ta1_result}, our DANML improves Recall@1 by $1.9\%$ on the CUB-200-2011, and $1.5\%$ on the Cars-196 over the recent state-of-the-art multi-similarity loss. This may be because the logistic loss function is more powerful than the linear function for generalization. Meanwhile, for recently proposed method Circle Loss, our DANML outperforms it about $0.9\%$ on the CUB-200-2011 and $2.2\%$ on the Cars-196 dataset. Compared with ABE which is an ensemble method with a much heavier model, our method achieves a higher Recall@1 by $7.0\%$ improvement on the CUB-200-2011 and $0.4\%$ on the Cars-196 dataset.\\
\indent For the Stanford Online Products (SOP) and the In-Shop Clothes Retrieval (In-Shop), as seen from Tables \ref{ta2_result} and \ref{ta3_result}, our method outperforms multi-similarity loss by $1.7\%$ on the In-Shop dataset and by $0.4\%$ on the SOP dataset, respectively. Furthermore, when compared with ABE, our method increases Recall@1 by $3.6\%$ and $2.8\%$ on the In-Shop and SOP dataset, respectively. For the Circle Loss which is a recent state-of-the-art method on SOP dataset, our DANML achieves a better performance about $1.6\%$ on it.\\
\indent To summarize, our method achieves new state-of-the-art or comparable performance on four data sets, even taking those methods with ensemble techniques like ABE and BIER into consideration. Those results also validate that it is very important to remove the inseparable samples for metric learning training.
\begin{table*}[!htbp]
\vspace{-0pt}\renewcommand{\arraystretch}{0.3} % 行间距
\setlength{\tabcolsep}{7pt}  % 列间距
\centering
\caption{Recall@K(\%) performance on CUB-200-2011 dataset and Cars-196 dataset. Superscript denotes embedding size.}
\label{ta1_result}
\footnotesize
\vspace{-6pt}\begin{tabular}{l|c|cccccc|cccccc}
\toprule
\multicolumn{1}{c|}{}&\multicolumn{1}{c|}{Publication}&\multicolumn{6}{c|}{CUB-200-2011}&\multicolumn{6}{c}{Cars-196}\\
\midrule
\multicolumn{2}{c|}{$Recall@K(\%)$}&1&2&4&8&16&32&1&2&4&8&16&32\\
\midrule
Clustering$^{64}$\cite{oh2017deep}&CVPR17&48.2&61.4&71.8&81.9&-&-&58.1&70.6&80.3&87.8&-&-\\
ProxyNCA$^{64}$\cite{movshovitz2017no}&ICCV17&49.2&61.9&67.9&72.4&-&-&73.2&82.4&86.4&87.8&-&-\\
Smart Mining$^{64}$\cite{harwood2017smart}&CVPR17&49.8&62.3&74.1&83.3&-&-&64.7&76.2&84.2&90.2&-&-\\
Margin$^{128}$\cite{wu2017sampling}&ICCV17&63.6&74.4&83.1&90.0&94.2&-&79.6&86.5& 91.9& 95.1& 97.3& -\\
HDC$^{384}$\cite{wu2017sampling}&CVPR17&53.6&65.7&77.0&85.6&91.5&95.5&73.7&83.2&89.5&93.8&96.7&98.4\\
HTL$^{512}$\cite{ge2018deep}&ECCV18&57.1&68.8&78.7&86.5&92.5&95.5&81.4&88.0&92.7&95.7&97.4&99.0\\
ABIER$^{512}$\cite{opitz2018deep}&PAMI18&57.5&68.7&78.3&86.2&91.9&95.5&82.0&89.0&93.2&96.1&97.8&98.7\\
ABE$^{512}$\cite{kim2018attention}&ECCV18&60.6&71.5&79.8&87.4&-& -&85.2&90.5&94.0&96.1&-&-\\
Multi-similarity loss$^{512}$\cite{wang2019multi}&CVPR19&65.7&77.0&86.3&\textbf{91.2}&95.0&97.3&84.1&90.4&94.0&96.5&98.0&98.9\\
Hardness-aware$^{512}$\cite{zheng2019hardness}&CVPR19&53.7&65.7&76.7&85.7&-&-&79.1&87.1&92.1&95.6&-&-\\
Circle Loss$^{512}$\cite{wang2020ranked}&CVPR20&66.7&77.4&86.2&\textbf{91.2}&-&-&83.4&89.8&\textbf{94.1}&96.5&-&-\\
Ranked list loss$^{512}$\cite{wang2019ranked}&CVPR19&61.3&72.7&82.7&89.4&-&-&82.1&89.3&93.7&97.7&-&-\\
\midrule
%DANML$^{64}$&-&59.3&71.8&81.5&89.4&95.3&97.5&78.9&87.7&92.4&96.2&97.9&98.8\\
DANML$^{512}$&-&\textbf{67.6}&\textbf{79.1}&\textbf{86.4}&\textbf{91.2}&\textbf{97.1}&\textbf{98.1}&\textbf{85.6}&\textbf{92.1}&\textbf{94.1}&\textbf{97.7}&\textbf{98.1}&\textbf{99.3}\\
\bottomrule
\end{tabular}
\end{table*}
\begin{table}[!htbp]
\vspace{-0pt}\renewcommand{\arraystretch}{0.5} % 行间距
\setlength{\tabcolsep}{4pt}  % 列间距
\centering
\caption{Recall@K(\%) performance on In-Shop dataset. Superscript denotes embedding size.}
\label{ta2_result}
\footnotesize
\vspace{-6pt}\begin{tabular}{l|c|cccccc}
\toprule
\multicolumn{1}{c|}{}&\multicolumn{1}{c|}{Publication}&\multicolumn{6}{c}{In-Shop}\\
\midrule
\multicolumn{2}{c|}{$Recall @ k$}&1&10&20&30&40&50\\
\midrule
FashionNet$^{4096}$\cite{oh2017deep}&CVPR17&53.0&73.0&76.0&77.0&79.0&80.0\\
HDC$^{384}$\cite{wu2017sampling}&CVPR17&62.1&84.9&89.0&91.2&92.3&93.1\\
HTL$^{512}$\cite{ge2018deep}&ECCV18&80.9&94.3&95.8&97.2&97.4&97.8\\
ABIER$^{512}$\cite{opitz2018deep}&PAMI18&83.1&95.1&96.9&97.5&97.8&98.0\\
ABE$^{512}$\cite{kim2018attention}&ECCV18&87.3&96.7&97.9&98.2&98.5&98.7\\
\tiny{Multi-similarity loss}$^{512}$\cite{wang2019multi}&CVPR19&89.7&97.9&98.5&98.8&99.1&99.2\\
\midrule
%DANML$^{128}$&-&89.3&97.4&98.3&98.5&98.9&98.9\\
DANML$^{512}$&-&\textbf{90.1}&\textbf{98.2}&\textbf{98.9}&\textbf{99.0}&\textbf{99.3}&\textbf{99.4}\\
\bottomrule
\end{tabular}
\end{table}
\begin{table}[!htbp]
\vspace{-0pt}\renewcommand{\arraystretch}{0.5} % 行间距
\setlength{\tabcolsep}{5pt}  % 列间距
\centering
\caption{Recall@K(\%) performance on  SOP dataset. Superscript denotes embedding size.}
\label{ta3_result}
\footnotesize
\vspace{-6pt}\begin{tabular}{l|c|cccccc}
\toprule
\multicolumn{1}{c|}{}&\multicolumn{1}{c|}{Publication}&\multicolumn{3}{c}{SOP}\\
\midrule
\multicolumn{2}{c|}{$Recall @ k$}&1&10&100&1000\\
\midrule
Clustering$^{64}$\cite{oh2017deep}&CVPR17&67.0&83.7&93.2&-\\
ProxyNCA$^{64}$\cite{movshovitz2017no}&ICCV17&73.7&-&-&-\\
Smart Mining$^{64}$\cite{harwood2017smart}&CVPR17&49.8&62.3&74.1&-\\
Margin$^{38}$\cite{wu2017sampling}&ICCV17&72.7&86.2&93.8&98.0\\
HDC$^{384}$\cite{wu2017sampling}&CVPR17&69.5&84.4&92.8&97.7\\
HTL$^{512}$\cite{ge2018deep}&ECCV18&74.8&88.3&94.8&98.4\\
ABIER$^{512}$\cite{opitz2018deep}&PAMI18&74.2&86.9&94.0&97.8\\
ABE$^{512}$\cite{kim2018attention}&ECCV18&76.3&88.4&94.8&98.2\\
Multi-similarity loss$^{512}$\cite{wang2019multi}&CVPR19&78.2&90.5&96.0&98.7\\
Hardness-aware$^{512}$\cite{zheng2019hardness}&CVPR19&68.4&83.5&92.3&-\\
Ranked list loss$^{512}$\cite{wang2019ranked}&CVPR19&79.8&91.3&96.3&-\\
Circle Loss$^{512}$\cite{wang2020ranked}&CVPR20&78.3&90.5&96.1&98.6\\
\midrule
%DANML$^{64}$&-&77.1&90.0&95.9&98.8\\
DANML$^{512}$&-&\textbf{79.9}&\textbf{92.1}&\textbf{96.4}&\textbf{98.9}\\
\bottomrule
\end{tabular}
\end{table}
\section{Conclusion}
In this paper, we proposed a general framework named adaptive neighborhood metric learning (ANML) to solve the inseparable problem of metric learning, which has very interesting properties. That is by setting its parameters different values, the proposed method can be seen as the improvements of the existing state-of-the-art methods. For learning linear projection, we prove that LMNN and NCA are the special cases of LANML. Compared with the original versions of LMNN and NCA, the LANML has a boarder searching space and may have more appropriate solutions. When we select the deep neural network as projection function, our model can be seen as the improvement of existing methods, such as the N-pairs loss, Proxy-NCA, and multi-similarity loss. We evaluate our algorithms on the UCI datasets and the large scale image datasets. The promising results show the superiority of the proposed method.

%\ifCLASSOPTIONcompsoc
%  % The Computer Society usually uses the plural form
\section{Acknowledgments}
{This work was supported in part by the National Key R\&D Program of China under Grant 2017YFB1002201, and by the National Natural Science Foundation of China under Grant 61772425 and 61822603.}
%\else
%  % regular IEEE prefers the singular form
%\fi

\ifCLASSOPTIONcaptionsoff
  \newpage
\fi



%\bibliographystyle{IEEEtran}
\vspace{-10pt}\bibliographystyle{ieeetr}
\bibliography{bare_jrnl}
\ifCLASSOPTIONcaptionsoff
  \newpage
\fi
\begin{IEEEbiography}[{\includegraphics[width=1in,height=1.25in,clip,keepaspectratio]{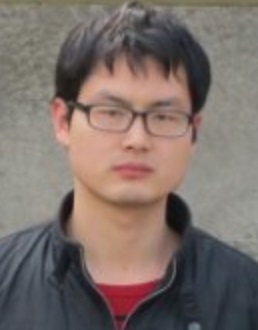}}]{Kun Song}
received his master degree and Ph.D degree from the Northwestern Polytechnical University, Xi'an, China, in 2015 and 2020, respectively. His research interests include computer vision and machine learning.
\end{IEEEbiography}

\begin{IEEEbiography}[{\includegraphics[width=1in,height=1.25in,clip,keepaspectratio]{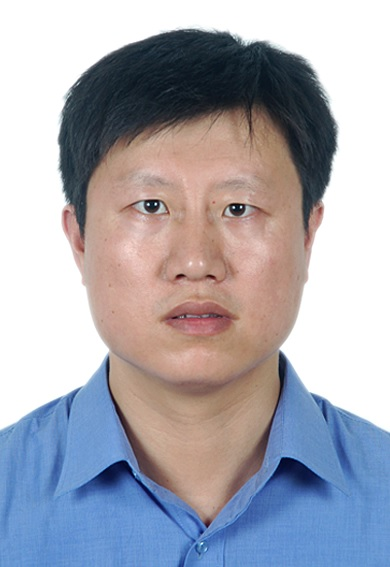}}]{Junwei Han}
(M'12 - SM'15) received the Ph.D degree in pattern recognition and intelligent systems from the School of Automation, Northwestern Polytechnical University in 2003. He is a currently a Professor with Northwestern Polytechnical University, Xi'an, China. His research interests include multimedia processing and brain imaging analysis. He is an Associate Editor of the IEEE Transactions on Human-Machine Systems, Neurocomputing, Machine Vision and Applications, and Multidimensional Systems and Signal Processing.
\end{IEEEbiography}

\begin{IEEEbiography}[{\includegraphics[width=1in,height=1.25in,clip,keepaspectratio]{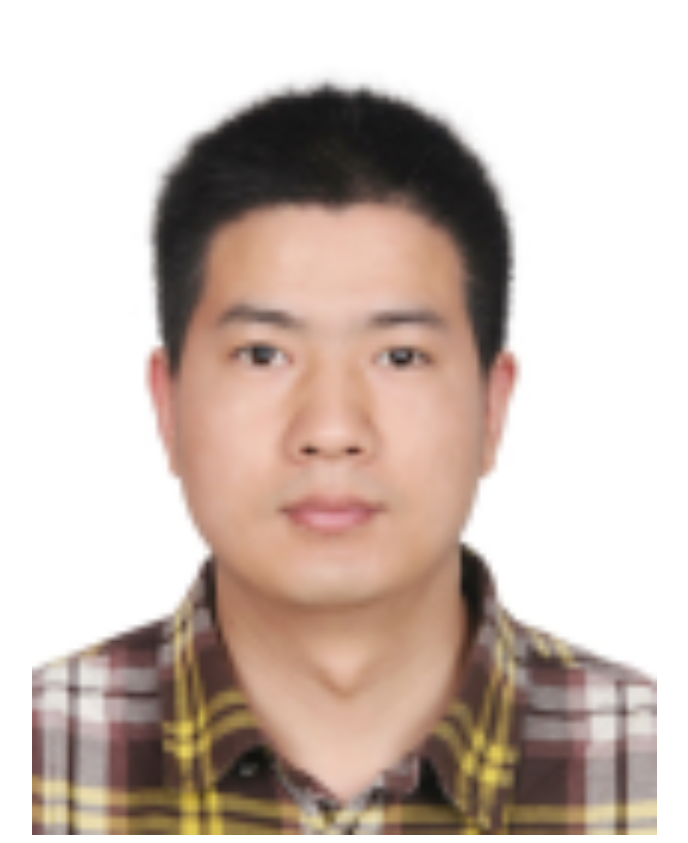}}]{Gong Cheng}
Gong Cheng received the B.S. degree from Xidian University, Xi'an, China, in 2007, and the M.S. and Ph.D. degrees from Northwestern Polytechnical University, Xi'an, China, in 2010 and 2013, respectively. He is currently a Professor with Northwestern Polytechnical University, Xi'an, China. His main research interests are computer vision and pattern recognition.
\end{IEEEbiography}
%
%\cite{}
%
%
%
%
\begin{IEEEbiography}[{\includegraphics[width=1in,height=1.25in,clip,keepaspectratio]{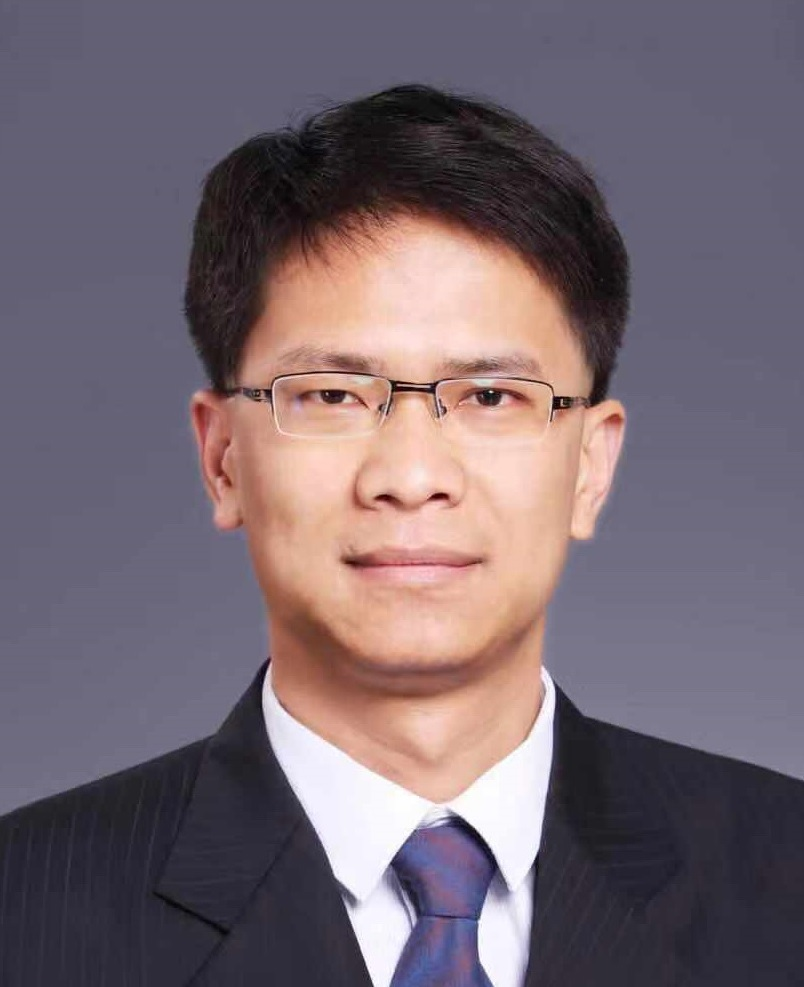}}]{Jiwen Lu} (M'11-SM'15) received the B.Eng. degree in mechanical engineering and the M.Eng. degree in electrical engineering from the Xi'an University of Technology, Xi'an, China, in 2003 and 2006, respectively, and the Ph.D. degree in electrical engineering from Nanyang Technological University, Singapore, in 2012. He is currently an Associate Professor with the Department of Automation, Tsinghua University, Beijing, China. His current research interests include computer vision and pattern recognition. He serves the Co-Editor-of-Chief of the Pattern Recognition Letters, an Associate Editor of the IEEE Transactions on Image Processing, the IEEE Transactions on Circuits and Systems for Video Technology, the IEEE Transactions on Biometrics, Behavior, and Identity Science, and Pattern Recognition. He was/is a member of the Image, Video and Multidimensional Signal Processing Technical Committee, Multimedia Signal Processing Technical Committee, and the Information Forensics and Security Technical Committee of the IEEE Signal Processing Society, and a member of the Multimedia Systems and Applications Technical Committee and the Visual Signal Processing and Communications Technical Committee of the IEEE Circuits and Systems Society. He is a senior member of the IEEE and an IAPR Fellow.
\end{IEEEbiography}

\begin{IEEEbiography}[{\includegraphics[width=1in,height=1.25in,clip,keepaspectratio]{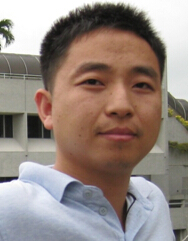}}]{Fieping Nie}
received the Ph.D. degree in computer Science from Tsinghua University, China in 2009. His research interests are machine learning and its applications, such as pattern recognition, data mining, computer vision, image processing and information retrieval. He has published more than 100 papers in the following top journals and conferences: TPAMI, IJCV, TIP, TNNLS/TNN, TKDE, TKDD, Bioinformatics, ICML, NIPS, KDD, IJCAI, AAAI, ICCV, CVPR. His papers have been cited more than 15000 times (Google scholar). He is now serving as Associate Editor or PC member for several prestigious journals and conferences in the related fields.
\end{IEEEbiography}
\end{document}